\documentclass[ijoc,sglanonrev]{informs4}
\usepackage{eqndefns-left} % For checking the display equation width and equation environment definitions %
\RequirePackage{tgtermes}
\RequirePackage{newtxtext}
\RequirePackage{newtxmath}
\RequirePackage{bm}
\RequirePackage{endnotes}

\OneAndAHalfSpacedXII % Current default line spacing
\usepackage{algorithm}
\usepackage{algpseudocode}
\usepackage{tikz}

\usepackage{natbib}
 \bibpunct[, ]{(}{)}{,}{a}{}{,}%
 \def\bibfont{\small}%
\EquationsNumberedThrough    % Default: (1), (2), ...
\TheoremsNumberedThrough     % Preferred (Theorem 1, Lemma 1, Theorem 2)
\ECRepeatTheorems  %  

\MANUSCRIPTNO{}

\usepackage{amsmath,amsfonts}
\usepackage{array}
\usepackage[caption=false,font=normalsize,labelfont=sf,textfont=sf]{subfig}
\usepackage{textcomp}
\usepackage{stfloats}
\usepackage{url}
\usepackage{verbatim}
\usepackage{graphicx}
\usepackage{xcolor} % text color
\usepackage{booktabs} % for better table lines
\usepackage{multirow} % For \multirow command

\algrenewcommand\algorithmicindent{0.6em}
\long\def\theARTICLETOPLEFT{}
\long\def\theARTICLETOPRIGHT{}

\RRHFirstLine{}\RRHSecondLine{}
\LRHFirstLine{}\LRHSecondLine{}

\makeatletter
\def\setoddRH{\hbox to \textwidth{\fs.7.8.\tabcolsep0pt
  \begin{tabular*}{\textwidth}[b]{l@{\extracolsep\fill}r}
  {\theRRHFirstLine}&\\ {\theRRHSecondLine}&\\[-4pt]
  \rlap{\VRHDW{0.5pt}{0pt}{\textwidth}}&\\
  \end{tabular*}}}
\def\setevenRH{\hbox to \textwidth{\fs.7.8.\tabcolsep0pt
  \begin{tabular*}{\textwidth}[b]{l@{\extracolsep\fill}r}
  &{\theLRHFirstLine}\\ &{\theLRHSecondLine}\\[-4pt]
  \rlap{\VRHDW{0.5pt}{0pt}{\textwidth}}&\\
  \end{tabular*}}}
\def\setoddRF{\vbox{\vskip25pt\hbox to \textwidth{\hfil\fs.8.8.\thepage\hfil}}}
\def\setevenRF{\vbox{\vskip25pt\hbox to \textwidth{\hfil\fs.8.8.\thepage\hfil}}}
\makeatother

\begin{document}
%%%%%%%%%%%%%%%%

%\setcounter{page}{1}
%\DOI{10.1287/xxxx.0000.0000}%

\RUNAUTHOR{Mingze, Juan and Jianbang}

% \RUNTITLE{Predictive Maintenance in Manufacturing}
\RUNTITLE{Diffuse to Detect: A Generalizable Framework for Anomaly Detection with Diffusion Models – Applications to UAVs and Beyond}

% \TITLE{Optimal Resource Allocation in Humanitarian Logistics: A Stochastic Programming Approach}

\TITLE{Diffuse to Detect: A Generalizable Framework for Anomaly Detection with Diffusion Models – Applications to UAVs and Beyond}

\ARTICLEAUTHORS{%
%\AUTHOR{John Doe,\textsuperscript{a} Jane Smith,\textsuperscript{b}}
%\AFF{\textsuperscript{a}Department of Industrial Engineering, University of XYZ, \EMAIL{john.doe@xyz.edu; \textsuperscript{b}Department of Computer Science, University of ABC, \EMAIL{jane.smith@abc.edu}} 
\AUTHOR{Mingze Gong\footnotemark[1]}
\AFF{
The Hong Kong University of Science and Technology (Guangzhou), \EMAIL{mgong081@connect.hkust-gz.edu.cn}}

\AUTHOR{Juan Du\footnotemark[1]}
\AFF{
The Hong Kong University of Science and Technology (Guangzhou), The Hong Kong University of Science and Technology, \EMAIL{juandu@ust.hk}}

\AUTHOR{Jianbang You}
\AFF{
	The Hong Kong University of Science and Technology (Guangzhou), The Hong Kong Polytechnic University, \EMAIL{jianbangy@hkust-gz.edu.cn}}
} % end of the block

\ABSTRACT{%
% Enter your abstract
Anomaly detection in complex, high-dimensional data, such as UAV sensor readings, is essential for operational safety but challenging for \textcolor{black}{existing} methods due to their limited sensitivity, scalability, and inability to capture intricate dependencies. We propose the Diffuse to Detect (DTD) framework, a novel approach that innovatively adapts diffusion models for anomaly detection, diverging from their conventional use in generative tasks with high inference time. By comparison, DTD employs a single-step diffusion process to predict noise patterns, enabling rapid and precise identification of anomalies without reconstruction errors. This approach is grounded in robust theoretical foundations that link noise prediction to the data distribution’s score function, ensuring reliable deviation detection. By integrating Graph Neural Networks to model sensor relationships as dynamic graphs, DTD effectively captures spatial (inter-sensor) and temporal anomalies. 
Its two-branch architecture, with parametric neural network-based energy scoring for scalability and nonparametric statistical methods for interpretability, provides flexible trade-offs between computational efficiency and transparency.
Extensive evaluations on UAV sensor data, multivariate time series, and images demonstrate DTD’s superior performance over existing methods, underscoring its generality across diverse data modalities. This versatility, combined with its adaptability, positions DTD as a transformative solution for safety-critical applications, including industrial monitoring and beyond.
}%

% \FUNDING{This research was supported by [grant number, funding agency].}

%Supplemental Material:
%Data Ethics & Reproducibility Note:

%\KEYWORDS{Stochastic programming, Decision support,Uncertainty, Disaster response, Optimization}

%\HISTORY{Received: Month DD, YYYY; Accepted: Month DD, YYYY; Published Online: Month DD, YYYY}

\maketitle
\footnotetext[1]{These authors contributed equally to this work.}
%%%%%%%%%%%%%%%%%%%%%%%%%%%%%%%%%%%%%%%%%%%%%%%%%%%%%%%%%%%%%%%%%%%%%%

% Text of your paper here

\section{Introduction}
Unmanned Aerial Vehicles (UAVs) have become indispensable across diverse applications, including aerial surveillance, package delivery, disaster response, and environmental monitoring, due to their flexibility, cost-effectiveness, and ability to operate in challenging environments~\citep{shakhatreh2019}. As UAV deployment grows, ensuring operational reliability and safety becomes paramount, especially in high-stakes contexts like urban air mobility and infrastructure inspection. A key aspect of this is monitoring sensor data (encompassing metrics such as altitude, speed, battery levels, and engine performance) to gain insights into a UAV's operational health. \textcolor{black}{High-dimensional anomaly detection is essential for identifying critical issues, such as sudden engine failures, security breaches, or environmental disruptions, could compromise mission success or lead to accidents~\citep{yang2024e}.}
% deng2024 reference omitted for reference reduction.

UAV sensor data, however, poses significant challenges for effective anomaly detection. These data are high-dimensional, collected by numerous sensors with high sampling rates, and exhibit complex multivariate dependencies that shift dynamically based on flight conditions~\citep{huang2025}. Anomalies, such as engine failures, often manifest as abrupt change in sensor readings (e.g., simultaneous drops in power and altitude). Furthermore, UAVs produce diverse data modalities, including structured time series, graphs of sensor interactions, and occasionally images, necessitating detection frameworks capable of handling heterogeneous inputs~\citep{wang2019b}. Conventional statistical and deep learning methods struggle to address these complexities, underscoring the need for tailored approaches to achieve rapid and reliable anomaly detection for UAV safety.

\subsection{Limitations of Existing Methods}
Current anomaly detection techniques, spanning statistical and deep learning paradigms, exhibit significant limitations when tasked with detecting anomalies in UAV sensor data. These limitations include:

\begin{itemize}
	\item \textbf{Difficulty Detecting Subtle and Abrupt Anomalies in High-Dimensional Data}: Methods like autoencoders or statistical tests often fail to detect anomalies in high-dimensional UAV data, where noise and complex patterns can obscure critical deviations. For instance, abrupt changes following an engine failure may resemble normal pattern locally, leading to low reconstruction errors or insufficient sensitivity, resulting in detection failures~\citep{huang2025}. % chen2024 omitted for reference reduction
	\item \textbf{Inadequate Modeling of Complex Sensor Dependencies}:  UAVs produce multivariate time series with intricate inter-sensor relationships. Traditional methods often treat sensors independently or use simplistic correlation models, resulting in detection failures of anomalies arising from
coordinated deviations, such as simultaneous drops in power and altitude post-engine failure, which are critical for accurate detection~\citep{lin2010}. % deng2024 omitted for reference reduction
	\item \textbf{Limited Use and Adaptation of Generative Models for Structured Data}: Generative models such as GANs and VAEs are predominantly applied to image-based anomaly detection, with limited use for time series or other structed data, as their high computational cost hinders anomaly detection in UAV operational monitoring, where such data is commonly collected. Their reliance on reconstruction errors often fails to capture complex anomalies,
	      %  like those induced by engine failures across multiple sensors,
	      and advanced generative models are rarely adapted for real-world engineering challenges involving noisy UAV data~\citep{liu2023g}. % ho2025 omitted for reference reduction
	\item \textbf{Trade-offs Between Scalability and Interpretability}: Neural network-based approaches scale well for large datasets but lack transparency, obscuring why anomalies are flagged. Conversely, interpretable methods like Kernel Density Estimation (KDE) struggle with high-dimensional data, hindering real-time detection in UAV monitoring. This trade-off complicates diagnosis in safety-critical scenarios~\citep{hu2020}. % rehman2024 omitted for reference reduction
	\item \textbf{Limited Generalization Across Diverse Data Modalities}: Most anomaly detection frameworks are designed for specific data types, such as time series or images, and struggle to generalize to others, such as graphs, without significant adjustments. This specialization hinders their application in real-world UAV systems, where heterogeneous data (e.g., sensor readings, images, network graphs) are collected, requiring multiple models and complicating deployment for comprehensive monitoring~\citep{wang2019b}. % huang2025 omitted
\end{itemize}

These challenges are intensified by the curse of dimensionality, where high-dimensional data becomes sparse \textcolor{black}{because the data points grow increasingly spread out and isolated across an exponentially expanding volume of space,}
 undermining distance-based or density-based detection methods~\citep{suboh2023}. % souiden2022 omitted.
Moreover, the rarity and diversity of anomalies in UAV operations, coupled with the scarcity of labeled data, necessitate unsupervised anomaly detection approaches~\citep{lin2024}. % liu2023h, yang2023e, omitted
However, developing robust unsupervised methods that can effectively handle the high-dimensional and noisy UAV sensor data remains a significant challenge.
Together, these limitations highlight the inadequacy of existing approaches for rapidly and reliably detecting critical anomalies in UAV operations without labeled data.
% particularly those as severe as engine failures.

\subsection{Contributions of DTD}
The identified limitations underscore the need for an anomaly detection framework that is sensitive, scalable, interpretable, and adaptable to the complex, multimodal data from UAVs. Diffusion models (DMs) predict noise patterns in a manner aligned with detecting anomalous deviations in UAV data, with a detailed theoretical derivation provided later. Inspired by recent advancements in generative modeling for anomaly detection~\citep{livernoche2023, wang2023e}, we propose the Diffuse to Detect (DTD) framework, a customized generative approach for rapid and effective anomaly detection. Unlike conventional methods relying on reconstruction errors to identify anomalies, DTD directly evaluates data deviations using the learned patterns from diffusion models, enhancing efficiency and supported by theoretical validation that validates its efficacy for anomaly detection. It provides two diffusion-based pathways: DM-P, employing parametric scoring for efficient large-scale data handling of large-scale datasets, and DM-NP, utilizing nonparametric scoring for transparent and interpretable analysis. By reducing inference time compared to traditional anomaly detection generative models based on reconstruction, DTD addresses the real-time demands of UAV monitoring requirements, thereby improving safety and reliability through the following contributions:\begin{itemize}
	\item  \textbf{Enhanced Sensitivity and Effective Modeling of Sensor Dependencies}: Our framework, leveraging the unique mechanism of diffusion models to predict noise patterns, effectively detects subtle irregularities in high-dimensional UAV sensor data, such as early mechanical faults. \textcolor{black}{Additionally, the DTD framework transforms UAV time series into graphs modeling relationships such as among altitude, speed, engine power, and other parameters. This approach captures inter-sensor and temporal dependencies, enabling nuanced anomaly detection.}
	% \item \textbf{Enhanced Sensitivity to Subtle Anomalies}: Our framework, thanks to its unique mechanism of diffusion models to predict noise patterns, effectively detects subtle irregularities in high-dimensional UAV sensor data. This enables the identification of critical issues like early mechanical faults or minor environmental disruptions, significantly improving UAV safety and reliability.
	% \item \textbf{Effective Modeling of Sensor Dependencies}: We address the gap in capturing complex spatial and temporal sensor dependencies by transforming UAV time series into graphs, with edges modeling relationships like altitude, speed, and engine power. Our framework captures both relational (spatial) and sequential (temporal) aspects through the transformation, which provides a more holistic view of the system, allowing a more nuanced detection of anomalies.
	\item \textbf{Advanced Application of Generative Models for Structured and Multimodal Data}: We pioneer the development of diffusion models for anomaly detection in structured UAV data, such as time series and graphs, overcoming the limitations of reconstruction-based methods like GANs and VAEs. This framework also generalizes to diverse data types, including images, ensuring robust monitoring for potential real-world UAV applications.
	\item  \textbf{Scalable and Interpretable Anomaly Detection Across Diverse Modalities}: Our dual-branch framework leverages diffusion models, with branches selected based on user requirements to address distinct scenarios. The parametric branch, using energy-based scoring, ensures efficient processing of large, high-dimensional UAV data, while the nonparametric branch (KNN, KDE, iForest) provides transparent, diagnostic scoring for smaller-scale scenarios.
    % to balance scalability and interpretability, offering a parametric branch with energy-based scoring for efficient processing of large, high-dimensional UAV data and a nonparametric branch (KNN, KDE, iForest) for transparent, diagnostic scoring in smaller-scale scenarios. 
    By extracting robust features from complex data, such as sensor graphs, and enabling robust anomaly detection across diverse modalities such as time series, graphs, and images, it provides reliable, interpretable metrics tied to deviations. This unified approach eliminates the need for separate models, enhancing versatility, trust, and deployment efficiency for safety-critical UAV applications~\citep{hu2020, deng2024}. %rehman2024 0907 omitted
	% \item \textbf{Flexible Scalability with Reliable and Interpretable Scoring}: Our dual-branch framework addresses scalability and interpretability trade-offs by offering a DM-based neural network-based parametric branch, using energy-based scoring for efficient processing of large, high-dimensional UAV data, and a DM-NP branch (KNN, KDE, iForest) for transparent, diagnostic scoring in smaller-scale scenarios. By extracting robust features from complex data, such as sensor graphs, the DM enhances anomaly separability, while energy-based parametric and nonparametric density scores provide reliable, interpretable metrics directly tied to deviations, like engine failure anomalies, improving trust and usability in safety-critical UAV operations~\citep{rehman2024, hu2020, deng2024}.
	% \item \textbf{Unified Anomaly Detection Across Diverse Modalities}: Our framework enables robust anomaly detection across time series, graphs, and images, streamlining monitoring of UAVs’ heterogeneous data, such as sensor readings and network graphs.
	%       % affected by engine failures
	%       By adapting a single generative modeling approach to multiple data types, it eliminates the need for separate models, enhancing versatility and deployment efficiency for real-world UAV applications.
\end{itemize}

% Our framework provides tailored solutions for detecting critical anomalies in UAV operations. By tackling high dimensionality, complex sensor dependencies, and Multimodal data with efficient and interpretable approaches, it enhances safety and reliability, setting the stage for advanced anomaly detection in next-generation UAV systems. The subsequent sections detail our methodology and validate its performance across diverse datasets, including experiments on engine failure detection.

\section{Related Works}

\textcolor{black}{Research on anomaly detection in high-dimensional sensor data is broad, spanning reconstruction models, probabilistic likelihood models, and graph-centric approaches. First, reconstruction objectives often miss subtle or abrupt anomalies that remain locally consistent with normal patterns. Second, distance and subspace scores deteriorate as dimensionality increases or as correlations evolve over time. Third, static graph formulations capture fixed structure but struggle to track the multivariate dependencies that change during flight~\citep[see, e.g.,][]{kim2023a, ben-gal2023, tao2025, xu2025}. With this context, we now turn to UAV-specific methods.
}

\subsection{Statistical Methods for UAV Anomaly Detection}\label{sec:stat}
Classical statistical methods, such as Gaussian Mixture Models (GMM), Support Vector Machines (SVM), and Principal Component Analysis (PCA), have been foundational for anomaly detection across domains~\citep{chen2024}. GMMs assume data follows a mixture of Gaussian distributions, but UAV sensor data often exhibits non-Gaussian characteristics, leading to suboptimal performance and sensitivity to outliers~\citep{huang2025}. %veracini2009 omitted 
The computational cost of GMMs also escalates in high-dimensional spaces due to increasing parameter estimation~\citep{yu2020}. SVMs, while effective for some datasets, are computationally intensive for large, high-dimensional UAV data and lack interpretability with non-linear kernels, complicating diagnosis in safety-critical applications. % ~\citep{rehman2024} 0907 omitted
PCA reduces dimensionality but assumes linear relationships, potentially discarding critical information for detecting subtle anomalies in non-linear UAV data~\citep{huang2025}. % ahmed2024 omitted
These methods struggle with the curse of dimensionality, where sparse high-dimensional spaces render distance-based metrics less effective, reducing their ability to detect nuanced deviations~\citep{suboh2023}. % souiden2022 omitted

\subsection{Deep Learning Methods for UAV Anomaly Detection}\label{sec:dl}
Deep learning approaches, particularly for time series data, include Autoencoders (AEs), Recurrent Neural Networks (RNNs), and their variants like LSTMs and GRUs. % ozkat2024 omitted ~\citep{rehman2024} 0907
AEs detect anomalies via high reconstruction errors, but they may reconstruct subtle anomalies with low error, leading to missed detections~\citep{dhakal2023a}. RNNs capture temporal dependencies but struggle with complex multivariate interactions in high-dimensional UAV data~\citep{zhou2024a}. Generative models like GANs and VAEs, while effective for image anomaly detection, are underutilized for structured data like time series or graphs. 
When applied, these models rely on reconstruction errors, limiting sensitivity to anomalies that align locally with normal patterns \citep{ho2025, liu2023g}. % niu2020 omitted 
Advanced generative models, such as diffusion models, remain underexplored for UAV anomaly detection, representing an opportunity for capturing complex data distributions~\citep{zhang2024b}.

% \subsection{Anomaly Detection in High-dimensional Sensor data}\label{sec:highdim}
% Research on anomaly detection in high-dimensional sensor data is broad, spanning reconstruction models, probabilistic likelihood models, and graph-centric approaches. Rather than attempting a comprehensive survey, we highlight literatures that matter for UAVs. First, reconstruction objectives often miss subtle or abrupt anomalies that remain locally consistent with normal patterns. Second, distance and subspace scores deteriorate as dimensionality increases or as correlations evolve over time. Third, static graph formulations capture fixed structure but struggle to track the multivariate dependencies that change during flight~\citep[see, e.g.,][]{kim2023a, ben-gal2023, tao2025, xu2025}.

% These observations motivate predicting noise rather than reconstructing signals. A single-step diffusion model that predicts perturbations provides a scaled estimate of the normal-data score and produces sensitive anomaly signals without relying on high-dimensional reconstructions. 
% We adapt the idea to UAV telemetry and beyond, which are inherently multivariate, prioritizing sensitivity, scalability, and interpretability, while also establishing the soundness of the formulation through theoretical analysis.

\textcolor{black}{Overall, the literature reveals that prior work often misses non-linear inter-sensor relations, including altitude, groundspeed, and angular velocity, which leads to undetected interaction anomalies \citep{deng2024,lin2010}. High dimensionality degrades clustering and outlier methods \citep{suboh2023}, while real time use is limited by the scalability–interpretability trade-off \citep{hu2020} and by the need for separate models across modalities \citep{wang2019b}. % trade-off: rehman2024 0907 omitted
These gaps call for a unified, sensitive, and interpretable approach.
To address these gaps, our framework leverages diffusion models and Graph Neural Networks (GNNs) to capture non-linear dependencies, and by using single step noise prediction to retain sensitivity in high dimensions. Its two branch design balances scalable energy-based scoring (parametric) with interpretable diagnostics (nonparametric). The same framework applies to time series, graphs, and images, simplifying deployment while improving sensitivity, scalability, and interpretability for UAV anomaly detection tasks and broader applications.}

\section{Methodology}
\label{sec:methodology}

\subsection{Preliminaries}

\begin{figure*}[h]
	\centering
	\includegraphics[width=0.99\textwidth]{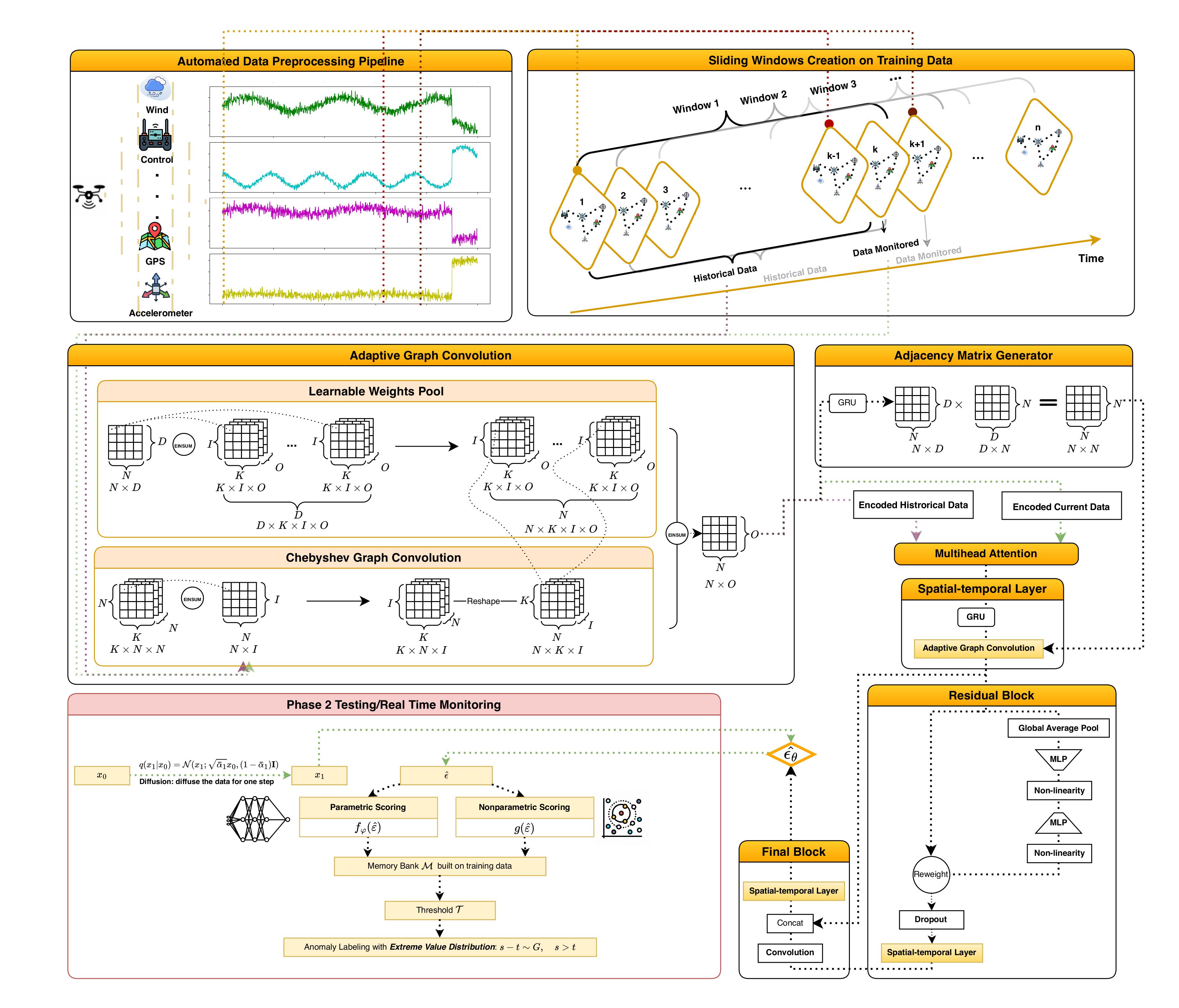}
	\caption{An Overview of the Framework. The Diffuse to Detect (DTD) framework first transforms UAV sensor data into a graph structure, where nodes represent sensors and edges initialized first and learnt through stochastic gradient descent. The framework then trains a diffusion model to learn the anomaly-free data distribution, enabling the prediction of noise patterns. For anomaly detection, the framework applies a single diffusion step to a test sample, generating a perturbed sample and predicting its noise. Two scoring branches (parametric and nonparametric) evaluate the predicted noise against the learned distribution, providing robust anomaly scores. Finally, the framework uses Extreme Value Theory (EVT) to compare scores against a threshold, enabling the detection of anomalies.}
	\label{fig:framework}
\end{figure*}

% Anomaly detection in UAVs involves identifying deviations in sensor data that indicate critical faults, such as engine failures, which can compromise safety and mission success. UAVs generate high-dimensional, multivariate time-series data from diverse sensors (e.g., altimeters, accelerometers, GPS).
% % , capturing metrics like altitude, speed, and engine power. 
% These sensors form a network of interdependent nodes, where anomalies often manifest as coordinated deviations across multiple signals, necessitating models that capture both inter-sensor and temporal dependencies~\citep{deng2024, huang2025}.
Anomaly detection in UAVs identifies deviations in sensor streams that signal critical faults such as engine failures. UAVs produce high dimensional, multivariate time series from heterogeneous sensors such as altimeter, accelerometer and GPS. These sensors form an interdependent network, anomalies often appear as coordinated shifts across signals, necessitating models that capture both inter sensor and temporal dependencies \citep{deng2024, huang2025}.

% To model these relationships, we represent UAV sensor data as a graph, where nodes correspond to sensors (or logical sensor groups, e.g., flight dynamics data combining altitude, speed, and throttle) derived from UAV telemetry. Edges encode dependencies, such as correlations between altitude and engine power. GNNs are well-suited for this task, as they learn representations by aggregating features from neighboring nodes, effectively capturing complex inter-sensor interactions~\citep{ma2023a}.
% In our framework, GNNs process graph-structured sensor data to extract spatial (inter-sensor) and temporal features, which are then used for anomaly detection. This approach addresses the limitations of traditional methods that treat sensors independently, which fails to detect critical anomalies arising from multivariate dependencies.
% % ~\citep{lin2010, chen2024}
% To model these relationships, we represent UAV sensor data as a graph, where nodes are sensors or logical groups (flight dynamics: altitude, speed, throttle), and edges encode dependencies such as the relation between altitude and engine power. 
\textcolor{black}{Inspired by GNNs that can aggregate neighbor information to capture inter-sensor interactions~\citep{ma2023a}, we model UAV sensor data relationships via a graph. The nodes, representing individual sensors or logical groups (flight dynamics: altitude, speed, throttle), provide spatial features, which enhance anomaly detection by overcoming independent sensor methods that miss correlated faults. The edges encode dependencies such as the relation between altitude and engine power.}

Formally, let \( \mathcal{G} = (\mathcal{V}, \mathcal{E}) \) denote the sensor graph, where \( \mathcal{V} = \{v_1, \dots, v_N\} \) represents \( N \) sensor nodes, each corresponding to a unique sensor, and \( \mathcal{E} \subseteq \mathcal{V} \times \mathcal{V} \) encodes edges based on domain knowledge or correlation analysis of sensor interactions. Each node \( v_i \), representing the \( i \)-th sensor, is associated with a feature vector \textcolor{black}{space} \( x_i(t) \in \mathbb{R}^d \) at time \( t \), capturing sensor measurements (e.g., climb rate, angular velocity).
As illustrated in the upper panel of Figure~\ref{fig:framework}, the multivariate time-series data from sensors is segmented using a sliding window approach, resulting in a dynamic graph sequence. 
The graph evolves over time, forming a sequence \( \{\mathcal{G}_t\} \), where each graph \( \mathcal{G}_t \) captures the sensor interactions and states at time \( t \).
% Anomaly detection aims to identify time points where the data distribution deviates significantly from the normal operational state, defined by the data distribution \( p_{\text{data}} \).
Each sensor \( v_i \) has a anomaly-free distribution \( p_i \), characterizing the normal behavior of its feature vector \( x_i(t) \). The collection of distributions, denoted \( p_{\text{data}} = \{p_i\}_{i=1}^N \), represents the expected behavior of the sensor network under normal conditions. 
Anomalies are defined according to sensor states where the observed data distribution of the \( i \)-th sensor deviates significantly from its normal distribution \( p_i \).
The final goal of anomaly detection is to output a binary label \(y^t \in \{0,1\}\) indicating whether the UAV given at time \(t\) is anomalous (\(1\)) or normal (\(0\)), given all sensor measurements.
To prevent confusion with the diffusion process in later sections, the sensor-specific index \( i \) is omitted when referring to general features, using \( x \) to denote a sample’s feature vector.

% Diffusion models (DMs) offer a powerful framework for this task by learning to model \( p_{\text{data}} \) through a noise-adding and denoising process. By integrating GNNs with DMs, our approach leverages graph-based feature extraction to enhance the sensitivity of diffusion-based anomaly detection, particularly for abrupt faults like engine failures. The following subsections detail how our DTD framework combines these elements to achieve robust, scalable, and interpretable anomaly detection for UAV sensor data.
% Diffusion models learn \(p_{\text{data}}\) through noise addition and denoising. We pair them with GNNs so graph-aware features guide the diffusion score, improving sensitivity to abrupt faults such as engine failures. The next subsections describe how DTD combines these elements to deliver robust, scalable, and interpretable anomaly detection for UAV sensor data.

\subsection{Overview of the DTD Framework}
% Diffusion models provide a principled way to capture the probabilistic structure of data. We leverage this for anomaly detection in UAV sensor streams. Given raw \textcolor{black}{anomaly-free} samples \(x_0 \sim p_{\text{data}}\), we train a network to predict the Gaussian perturbation \(\epsilon_\theta(x_k,k)\) added to a noisy sample \(x_k\) at diffusion step \(k \in \{0,\dots,T-1\}\) (with \(k=0\) denoting the clean sample).

Diffusion models provide a principled framework for modeling the probabilistic structure of data distributions.
% , which \textcolor{black}{inspired us to adapt diffusion model for} anomaly detection, enabling the identification of irregularities such as faults in UAV sensor measurements. 
These models learn an approximation \( p_\theta \) to the normal data distribution \( p_{\text{data}} \), where \( p_{\text{data}} \) represents the distribution of the original, noise-free (raw) data samples \( x_0 \sim p_{\text{data}} \).
We train a neural network that outputs \(\epsilon_\theta(x_k,k)\), the predicted Gaussian perturbation (\(\sim\mathcal{N}(0,I)\)) applied to a noisy sample \(x_k\) at diffusion step \(k\in\{0,\dots,T-1\}\). Here, \( k = 0 \) corresponds to the original, noise-free data, while larger values of \( k \) represent increasingly noisy versions of the sample.
% We train a neural network \textcolor{black}{and get output $\epsilon_\theta$} to predict the Gaussian noise component, drawn from $\mathcal{N}(0, I)$ and introduced to a perturbed sample $x_k$ at diffusion time step $k \in \{0, 1, \dots, T-1\}$. Here, \( k = 0 \) corresponds to the original, noise-free data, while larger values of \( k \) represent increasingly noisy versions of the sample.
Note that the diffusion time step \( k \) used in \( x_k \) refers to the level of noise applied during the diffusion process and should not be confused with the temporal index \( t \) in the graph sequence \( \{\mathcal{G}_t\} \), which denotes actual time steps in the sensor data. The diffusion time step \( k \) governs the denoising schedule, while the chronological order of UAV observations corresponds to the temporal sequence of sensor data.
Through training the neural network, the output $\epsilon_\theta$ captures the noise patterns characteristic of normal data under varying perturbation levels, effectively encoding the probabilistic structure of $p_{\text{data}}$ in its predictions.

% For anomaly detection in UAVs, as shown in the bottom part of Figure~\ref{fig:framework}, we apply a single diffusion step, at $k=1$, to a test sample $x$, yielding a lightly perturbed sample $x_1$. The trained model $\epsilon_\theta$ predicts the noise component of $x_1$, denoted $\hat{\epsilon}$. For normal data, $\hat{\epsilon}$ closely resembles the expected Gaussian noise distribution $\mathcal{N}(0, I)$, consistent with the model’s training on $p_{\text{data}}$. Anomalous data, such as erratic UAV sensor readings, diverge from $p_{\text{data}}$, leading to a predicted noise $\hat{\epsilon}$ that deviates from the expected distribution. \textcolor{black}{By limiting perturbation to a single step, our method preserves the intrinsic features of \( x \), enhancing sensitivity to subtle anomalies, such as minor sensor discrepancies. This approach outperforms methods relying on extensive noise accumulation or complete data reconstruction.}
For UAV anomaly detection (Fig.~\ref{fig:framework}), we apply a single forward diffusion step (\(k=1\)) to a test sample \(x\) to obtain \(x_1\). The network predicts the perturbation \(\hat{\epsilon}=\epsilon_\theta(x_1,1)\). For normal data, \(\hat{\epsilon}\) is approximately distributed as \(\mathcal{N}(0,I)\), while anomalies will show systematic deviations. By limiting perturbation to a single step, our method preserves the intrinsic features of \( x \), enhancing sensitivity to subtle anomalies, such as minor sensor discrepancies. 
% To quantify these deviations, we propose two scoring mechanisms that leverage the \textcolor{black}{learned representation reflected in the predicted noise \(\hat{\epsilon}\), each addressing different practical considerations for anomaly detection.}
% % , reflected in the predicted noise $\hat{\epsilon}$, addressing distinct practical considerations for anomaly detection. 
% The \textbf{parametric scoring} branch, selected for scenarios demanding scalability for high-dimensional UAV data, employs an energy-based model to compute anomaly scores based on the divergence of $\hat{\epsilon}$ from $\mathcal{N}(0, I)$. 
% The \textbf{nonparametric scoring} branch, selected for scenarios requiring transparent analysis and interpretability, applies statistical methods, such as k-nearest neighbors or kernel density estimation, to compare $\hat{\epsilon}$ against the noise distribution of normal data. 
% Each branch is independently selected \textcolor{black}{according to different scenarios}, leveraging the correspondence between the learned distribution $p_\theta$ and the noise characteristics modeled by $\epsilon_\theta$ to enable robust detection of anomalies, from abrupt faults to nuanced irregularities.
% This framework is particularly effective for real-time UAV monitoring, where computational efficiency and high precision are critical. The subsequent subsections provide a rigorous formulation, theoretical justification, and detailed scoring methodology, building on the foundation of diffusion model~\citep{ho2020}.
To quantify such deviations, we propose two scoring mechanisms that leverage the learned representation reflected in the predicted noise \(\hat{\epsilon}\), each addressing different practical considerations for anomaly detection.
The \textbf{parametric} branch scales to high-dimensional UAV data using an energy-based model that scores divergence from \(\mathcal{N}(0,I)\). 
The \textbf{nonparametric} branch favors interpretability, using kNN or KDE to compare \(\hat{\epsilon}\) against the normal noise distribution. 
Users select a branch based on scalability versus transparency, leveraging the link between \(p_\theta\) and the noise modeled by \(\epsilon_\theta\) to detect both abrupt faults and subtle irregularities. This design suits real-time UAV monitoring where both efficiency and precision are critical. The next sections provide the model formulation, theoretical justifications, and scoring details building on diffusion models \citep{ho2020}.

\subsection{Diffusion Model Formulation}

Diffusion models approximate the distribution $p_\theta$ as $p_{\text{data}}$ through a \textcolor{black}{Markov chain that corrupts data with incremental noise and learns the reverse transitions for denoising-based reconstruction}.
% , supporting the anomaly detection strategy outlined in the \textbf{Overview}. 
The \textbf{forward process} operates as a Markov chain, applying a fixed noise schedule to data, while the \textbf{reverse process} samples cleaner data using a trained neural network. 
% Below, we detail these processes, focusing on \textcolor{blue}{the} formulation and the training of the noise prediction model used for detecting anomalies in applications like UAV sensor data.
\textcolor{black}{Below, we clearly explain the training of noise prediction model used to detect anomalies and the formulation in UAV applications.}

The \textbf{forward process} perturbs a clean data sample $x_0 \in \mathbb{R}^d$, drawn from $p_{\text{data}}$, over $T$ diffusion time steps $k = 0, 1, \dots, T-1$. Gaussian noise is added incrementally, producing \textcolor{black}{noisy} samples $x_k$. Mathematically, at diffusion time step $k$:
\begin{equation}
	x_k = \sqrt{\bar{\alpha}_k} x_0 + \sqrt{1 - \bar{\alpha}_k} \epsilon, \quad \epsilon \sim \mathcal{N}(0, I),
	\label{eq:forward_diffusion}
\end{equation}
where $\bar{\alpha}_k = \prod_{s=0}^k \alpha_s$ with $\alpha_s = 1 - \beta_s$, $\beta_s \in (0, 1)$ a predefined noise increment, and $\epsilon$ is standard Gaussian noise. At $k=0$, $\bar{\alpha}_0 = 1$, so $x_0$ is unchanged; at $k=T-1$, $\bar{\alpha}_{T-1} \approx 0$, making $x_{T-1}$ nearly pure noise. This forward process is characterized by a conditional Gaussian distribution:
\begin{equation}
	q(x_k | x_0) = \mathcal{N}\left(x_k \Big| \sqrt{\bar{\alpha}_k} x_0, (1 - \bar{\alpha}_k) I\right),
	\label{eq:forward_prob}
\end{equation}
indicating that $x_k$ combines the original data $x_0$, scaled by $\sqrt{\bar{\alpha}_k}$, with noise of variance $1 - \bar{\alpha}_k$. In a UAV context, $x_k$ represents sensor data with increasing perturbations, similar to noise encountered during flight.

% In standard diffusion models, the \textbf{reverse process} approximates a Markov chain and seeks to reconstruct $x_0$ from a noisy $x_{T-1}$ by modeling the conditional distribution $p_\theta(x_{k-1} | x_k)$, approximated as a Gaussian:
% \begin{equation}
% 	p_\theta(x_{k-1} | x_k) = \mathcal{N}\left(x_{k-1} \Big| \mu_\theta(x_k, k), \beta_k I\right),
% 	\label{eq:reverse_prob}
% \end{equation}
% where the random variable $x_{k-1}$ follows a Gaussian distribution with mean $\mu_\theta(x_k, k)$ and variance $\beta_k I$, fixed for computational efficiency, as derived by~\cite{ho2020} via a noise prediction model $\epsilon_\theta(x_k, k)$ that estimates the noise $\epsilon$ in $x_k$. 
In generative tasks, the \textbf{reverse process} of diffusion models reconstructs clean data by iteratively denoising \( x_{T-1} \). However, its high computational cost, requiring thousands of steps, makes it unsuitable for real-time UAV anomaly detection. Our approach leverages only the \textbf{forward process}, directly analyzing the predicted noise for efficient anomaly detection, largely eliminating the need for iterative and time-consuming reconstruction. Further details on the \textbf{reverse process} and complete reconstruction of $x_0$ refer to~\cite{ho2020}.

% To support our anomaly detection framework, we utilize the noise prediction capability of diffusion models by training a neural network $\epsilon_\theta$ to estimate the noise $\epsilon$ introduced in \eqref{eq:forward_diffusion}, given the noisy sample $x_k$, its corresponding historical window $x_{hist}$, and diffusion time step $k$. 
To support our anomaly detection framework, we utilize the noise prediction capability of diffusion models. A neural network, parameterized by \(\theta\), is trained to predict the noise \(\epsilon_\theta\), approximating the noise \(\epsilon\) introduced in Equation ~\eqref{eq:forward_diffusion}, using the noisy sample \(x_k\), its historical window \(x_{hist}\), and diffusion time step \(k\).
The noise $\epsilon$ is intrinsically linked to the data distribution, as $x_k$ is generated from $x_0$ and $\epsilon$ according to Equation~\eqref{eq:forward_diffusion}, enabling $\epsilon_\theta$ to capture patterns of normal data. We leverage this predicted noise for anomaly detection, with theoretical justification provided in the next Subsection~\ref{noise_based_ad} and detailed derivations in Appendix~\ref{app:proof_noise_score}.

We completely redesigned the model to effectively process UAV data with inter-sensor and temporal dependencies, tailoring it for anomaly detection in UAV sensor systems.
Predefining sensor relationships is impractical in this context due to the large number of sensors on a UAV and their complex interactions. 

To address the challenge of modeling sensor relationships without relying on a predefined graph structure, we further developed the adaptive structure proposed by~\cite{bai2020} and implemented a purely data-driven approach to infer these relationships, as shown in Figure~\ref{fig:framework}. Our adaptive graph convolution module employs a learnable weights pool and Chebyshev graph convolution~\citep{he2022a} to capture spatial dependencies between sensors dynamically, eliminating the need for a static graph.
We fuse encoded history with the current test sample using multihead attention~\citep{NIPS2017_3f5ee243}, aligning long-range context with present observations to better expose subtle anomalies in multivariate UAV sensors. The fused features then pass to a spatiotemporal layer that combines a GRU for temporal dynamics with adaptive graph convolution for spatial dependencies, producing a comprehensive representation for detection.
% Our anomaly detection task processes both encoded historical data and current test data, which are fused using a multihead attention module. This fusion enhances the model’s ability to integrate long-range temporal dependencies and contextual relationships between historical patterns and current observations, significantly improving the detection of subtle anomalies in multivariate UAV sensor data. The fused output is then fed into a spatiotemporal layer, which combines a GRU to model temporal dynamics and the adaptive graph convolution module to capture spatial sensor dependencies. This dual mechanism ensures that both spatial and temporal patterns are comprehensively represented.
Finally, to produce noise predictions, we add a residual block followed by a projection block. The residual block stabilizes training and reduces overfitting, while the projection block concatenates features and applies a convolution to yield a refined representation for real-time parametric and nonparametric scoring.
% Finally, to generate noise predictions, we incorporate a residual block followed by a final block. The residual block facilitates effective training of the deep network by mitigating gradient issues and preventing overfitting. The final block integrates the processed features through concatenation and convolution, producing a refined representation that supports real-time anomaly detection via parametric and non-parametric scoring, as detailed in the framework.

Our framework includes two branches, each selected for specific scenarios: the parametric branch prioritizes computational efficiency, while the nonparametric branch emphasizes interpretability.
% \textcolor{red}{(please refer to the Scoring Methodology section for branch choice selection,)}
We employ a cooperative training strategy to fully leverage the model's capacity, as detailed in Algorithm~\ref{alg:training_dm}. 
Specifically, 
% for each training iteration, we sample a data point \( x_0 \) from the training dataset, select a random time step \( k \). Then draw noise \( \epsilon \) from a standard normal distribution \( \mathcal{N}(0, I) \), and compute the noisy sample \( x_k \) using the forward diffusion process. 
% for each training iteration, we sample a data point $x_0$ from the dataset and choose a random diffusion time step $k$. 
% We then draw Gaussian noise $\epsilon \sim \mathcal{N}(0,I)$ and obtain the noisy sample $x_k$ using the forward process.
at each iteration, we draw \(x_0\) from the dataset, pick a random diffusion step \(k\), sample \(\epsilon\sim\mathcal{N}(0,I)\), and form \(x_k\) via the forward process.

The diffusion loss, which forms the first part of the training objective, is calculated as:

\begin{equation}
	\mathcal{L}_{\text{DM}} = \mathbb{E}_{x_0 \sim p_{\text{data}}, \epsilon \sim \mathcal{N}(0, I), k} \left[ \|\epsilon - \epsilon_\theta(x_k, k, x_{hist})\|_2^2 \right].
	\label{eq:diffusion_loss}
\end{equation}

In addition to the diffusion loss \( L_{\text{DM}} \), each branch of the diffusion model incorporates a specific loss term to enhance anomaly detection capabilities.

For the nonparametric branch, an additional loss \( L_{\text{NP}} \) is computed using methods such as {Kernel Density Estimation (KDE)}, {k-Nearest Neighbors (kNN)}, or {Isolation Forest (iForest)}. This loss leverages a memory bank \( \mathcal{M} \) that stores predicted noise \( \hat{\epsilon}^+ \) from normal samples. The loss is calculated as:
\begin{equation}
	L_{\text{NP}} = \mathcal{L}_{\text{nonparam}}(\hat{\epsilon}^+, \hat{\epsilon}^-, \mathcal{M}),
\end{equation}
where:
\begin{itemize}
	\item \( \hat{\epsilon}^+ = \epsilon_\theta(x_0, 0, x_{\text{hist}}) \) is the predicted noise for a normal sample.
	\item \( \hat{\epsilon}^- = \epsilon_\theta(x_k, 0, x_{\text{hist}}) \) is the predicted noise for an anomalous or noisy sample.
	\item \( \mathcal{L}_{\text{nonparam}} \) distinguishes normal and anomalous noise patterns based on the chosen nonparametric method.
\end{itemize}
This approach allows the model to detect anomalies by comparing new noise predictions against a historical memory bank of normal patterns.

For the parametric branch, an {Energy-Based Model (EBM)} is employed, implemented as a neural network that processes predicted noise. The EBM uses positive samples \( \hat{\epsilon}^+ = \epsilon_\theta(x_0, 0, x_{\text{hist}}) \) from normal data and negative samples \( \hat{\epsilon}^- = \epsilon_\theta(x_k, 1, x_{\text{hist}}) \) from diffused data to distinguish between normal and anomalous patterns. The EBM loss is calculated as:
\begin{equation}
	L_{\text{EBM}} = E_\phi(\hat{\epsilon}^+) - E_\phi(\hat{\epsilon}^-).
\end{equation}

The total training loss combines the diffusion loss with the branch-specific loss, weighted by a hyperparameter \( \lambda \):
\begin{itemize}
	\item For the nonparametric branch:
	      \begin{equation}
		      L_{\text{total}} = L_{\text{DM}} + \lambda L_{\text{NP}}.
	      \end{equation}
	\item For the parametric branch:
	      \begin{equation}
		      L_{\text{total}} = L_{\text{DM}} + \lambda L_{\text{EBM}}.  
	      \end{equation}
\end{itemize}
% This cooperative training approach ensures that the model not only learns to accurately predict noise but also effectively distinguishes normal from anomalous patterns through the specialized mechanisms of each branch. The detailed scoring mechanisms for both branches are described in the later subsection.
This cooperative scheme learns accurate noise prediction and strong normal–anomaly discrimination. 
\textcolor{black}{Branch choice here depends on whether scalability and efficiency are prioritized with the parametric branch or interpretability and transparency with the nonparametric branch.
}
The following subsection details the scoring mechanisms.

% This whole strategy based on noise prediction is significantly faster than signal reconstruction methods (e.g., autoencoders) or the iterative \textbf{reverse process}, requiring only a single neural network evaluation. It is well-suited for UAV applications, where computational efficiency and rapid response are critical for safety. The theoretical underpinnings of why predicted noise effectively detects anomalies are explored in the next subsection.

\begingroup
\setlength{\abovecaptionskip}{1pt}
\setlength{\belowcaptionskip}{0pt}
\setlength{\textfloatsep}{3pt}
\setlength{\floatsep}{3pt}
\setlength{\intextsep}{3pt}
\begin{algorithm}[t]
\footnotesize
	\caption{Training Diffusion Model with Nonparametric or Parametric (EBM) Branch}
	\label{alg:training_dm}
	\begin{algorithmic}[1]
    \setlength{\itemsep}{0pt}\setlength{\parskip}{0pt}\setlength{\topsep}{0pt}
		\Require Dataset $D \sim p_{\text{data}}$ with sliding window pairs $(x_0, x_{\text{hist}})$, the pre-defined noise schedule $\{\beta_k\}_{k=0}^{T-1}$, a neural network $\epsilon_\theta(\cdot)$ parameterized by $\theta$, branch type
		\State Initialize model parameters of $\epsilon_\theta$
		\State Initialize empty memory bank: $\mathcal{M} \leftarrow \emptyset$
		\While{not converged}
		\State Sample $(x_0, x_{\text{hist}}) \sim D$
		\State Sample time step $k \sim \text{Uniform}\{0, 1, \ldots, T-1\}$
		\State Compute $\bar{\alpha}_k = \prod_{s=0}^k (1 - \beta_s)$
		\State Sample noise $\epsilon \sim \mathcal{N}(0, I)$
		\State Compute noisy sample: $x_k = \sqrt{\bar{\alpha}_k} x_0 + \sqrt{1 - \bar{\alpha}_k} \epsilon$
		\State Predict noise: $\hat{\epsilon} = \epsilon_\theta(x_k, k, x_{\text{hist}})$
		\State Compute diffusion loss: $L_{\text{DM}} = \left\| \epsilon - \hat{\epsilon} \right\|_2^2$

		\If{branch $==$ ``Nonparametric"} \Comment{e.g., KDE, kNN, iForest}
		\State Predict positive sample: $\hat{\epsilon}^+ = \epsilon_\theta(x_0, 0, x_{\text{hist}})$
		\State Predict negative sample: $\hat{\epsilon}^- = \epsilon_\theta(x_k, 0, x_{\text{hist}})$
		\State Compute nonparametric loss: $L_{\text{NP}} = \mathcal{L}_{\text{nonparam}}(\hat{\epsilon}^+, \hat{\epsilon}^-, \mathcal{M})$
		\If{memory bank is full}
		\State Remove oldest entry from $\mathcal{M}$ \Comment{FIFO update}
		\EndIf
		\State Update memory bank: $\mathcal{M} \leftarrow \mathcal{M} \cup \hat{\epsilon}^+$
		\State Compute total loss: $L_{\text{total}} = L_{\text{DM}} + \lambda L_{\text{NP}}$

		\ElsIf{branch $==$ ``Parametric"} \Comment{EBM-based refinement}
		\State Predict positive sample: $\hat{\epsilon}^+ = \epsilon_\theta(x_0, 0, x_{\text{hist}})$
		\State Predict initial negative sample: $\hat{\epsilon}^- = \epsilon_\theta(x_k, 0, x_{\text{hist}})$
		\State Refine negative sample via Langevin dynamics:
		\Statex \hspace{\algorithmicindent} $\hat{\epsilon}^- \leftarrow \text{Langevin}(\hat{\epsilon}^-, \nabla_{\hat{\epsilon}} E_\phi(\hat{\epsilon}))$
		\State Compute EBM loss: $L_{\text{EBM}} = E_\phi(\hat{\epsilon}^+) - E_\phi(\hat{\epsilon}^-)$
		\State Compute total loss: $L_{\text{total}} = L_{\text{DM}} + \lambda L_{\text{EBM}}$
		\EndIf

		\State Update model parameters: $\theta \leftarrow \text{Adam}(\theta, \nabla_\theta L_{\text{total}})$
		\EndWhile
		\State \textbf{Output:} Trained model $\epsilon_\theta$
	\end{algorithmic}
\end{algorithm}
\endgroup

\subsection{Noise-Based Anomaly Detection}\label{noise_based_ad}

To justify the efficacy of the predicted noise \(\epsilon_\theta(x_k, k, x_{hist})\) for anomaly detection, we establish Proposition~\ref{prop:noise_score} linking the noise prediction to the score function of the data distribution, offering a solid foundation for detecting deviations in UAV sensor data.
% The result is stated generally, with detailed derivations deferred to the appendix to maintain clarity in the main text.
\begin{proposition}
	\label{prop:noise_score}
	For a diffusion model with forward process \(x_k=\sqrt{\bar{\alpha}_k}\,x_0+\sqrt{1-\bar{\alpha}_k}\,\epsilon\), \(\epsilon\sim\mathcal{N}(0,I)\), the noise predictor satisfies
	\[
	\epsilon_\theta(x_k,k,x_{\text{hist}})\;\approx\;-\sqrt{1-\bar{\alpha}_k}\,\nabla_{x_k}\log p_k(x_k\mid x_{\text{hist}}),
	\]
	where \(p_k(x_k)\) is the distribution of \(x_k\). When \(k\) is small, such as \(k=1\), \(\epsilon_\theta(x_0, 1, x_{hist}) \approx -\sqrt{1 - \bar{\alpha}_0} \nabla_{x_0} \log p_{\text{data}}(x_0 | x_{hist})\) encodes the local geometry of \(p_{\text{data}}\).
	\end{proposition}

The derivation in Appendix~\ref{app:proof_noise_score} uses the forward process (Eq.~\ref{eq:forward_diffusion}) and Tweedie’s formula \citep{efron2011} to show that the optimal predictor \(\epsilon_\theta^*(x_k,k,x_{\text{hist}})=\mathbb{E}[\epsilon\mid x_k,x_{\text{hist}}]\) is proportional to the negative score \(\nabla_{x_k}\log p_k(x_k\mid x_{\text{hist}})\). The training objective drives \(\epsilon_\theta(x_k,k,x_{\text{hist}})\approx \epsilon_\theta^*\). Intuitively, for normal data, the prediction follows the noise pattern implied by \(p_{\text{data}}\) as the score points toward high-density regions, whereas faults induce deviations that \(\epsilon_\theta\) exposes as anomalies. 
Our approach's sensitivity and efficiency are supported by \(\epsilon_\theta(x_k, k, x_{hist})\), which captures distributional discrepancies to enable rapid and robust detection of subtle faults in UAVs, enhancing safety and reliability in critical scenarios.

\subsection{Scoring Methodology}

\begingroup
\setlength{\abovecaptionskip}{1pt}
\setlength{\belowcaptionskip}{0pt}
\setlength{\textfloatsep}{3pt}
\setlength{\floatsep}{3pt}
\setlength{\intextsep}{3pt}
\begin{algorithm}[t]
\footnotesize
	\caption{Testing Algorithm in  DTD Framework}
	\label{alg:testing_revised}
	\begin{algorithmic}[1]
    \setlength{\itemsep}{0pt}\setlength{\parskip}{0pt}\setlength{\topsep}{0pt}
		\Require Test sample $x$, trained diffusion model $\epsilon_\theta$, energy-based model $E_\phi$, memory bank $\mathcal{M} = \{\hat{\epsilon}_i^+\}_{i=1}^M$, time step $k=1$
		\State Sample noise $\epsilon \sim \mathcal{N}(0, I)$
		\State Compute $\bar{\alpha}_k = \prod_{s=0}^k (1 - \beta_s)$ for $k = 1$
		\State Compute noisy input: $x_k = \sqrt{\bar{\alpha}_k} x + \sqrt{1 - \bar{\alpha}_k} \epsilon$
		\State Predict noise: $\hat{\epsilon} = \epsilon_\theta(x_k, k, x_{hist})$
		\If{branch $==$ ``Nonparametric"}
			\State Compute anomaly score $s(x)$ using $\hat{\epsilon}$ and $\mathcal{M}$ \hfill \Comment{Higher score means anomalous}
		\ElsIf{branch $==$ ``Parametric"}
			\State Compute anomaly score: $s(x) = E_\phi(\hat{\epsilon})$ \hfill \Comment{Higher energy indicates anomalous}
		\EndIf
		\State \textbf{Output:} Anomaly score $s(x)$
	\end{algorithmic}
\end{algorithm}
\endgroup

% \begin{algorithm}
% 	\caption{Testing Algorithm in  DTD Framework}
% 	\label{alg:testing_revised}
% 	\begin{algorithmic}[1]
% 		\Require Test sample $x$, trained diffusion model $\epsilon_\theta$, energy-based model $E_\phi$, memory bank $\mathcal{M} = \{\hat{\epsilon}_i^+\}_{i=1}^M$, time step $k=1$
% 		\State Sample noise $\epsilon \sim \mathcal{N}(0, I)$
% 		\State Compute $\bar{\alpha}_k = \prod_{s=0}^k (1 - \beta_s)$ for $k = 1$
% 		\State Compute noisy input: $x_k = \sqrt{\bar{\alpha}_k} x + \sqrt{1 - \bar{\alpha}_k} \epsilon$
% 		\State Predict noise: $\hat{\epsilon} = \epsilon_\theta(x_k, k, x_{hist})$
% 		\If{branch $==$ ``Nonparametric"}
% 			\State Compute anomaly score $s(x)$ using $\hat{\epsilon}$ and $\mathcal{M}$ \hfill \Comment{Higher score = more anomalous}
% 		\ElsIf{branch $==$ ``Parametric"}
% 			\State Compute anomaly score: $s(x) = E_\phi(\hat{\epsilon})$ \hfill \Comment{Higher energy = more anomalous}
% 		\EndIf
% 		\State \textbf{Output:} Anomaly score $s(x)$
% 	\end{algorithmic}
% \end{algorithm}

The testing procedure for the DTD framework is outlined in Algorithm~\ref{alg:testing_revised}, which details the computation of anomaly scores for a test sample using the trained diffusion model and scoring branch. This algorithm leverages the predicted noise from a lightly perturbed input at diffusion time step $k=1$ to assess deviations from normal data distributions, enabling efficient and robust anomaly detection in UAV sensor data and beyond. Instead of directly predicting the noise at $k=0$, we apply a single diffusion step to the test sample. The justification for this approach is provided in Appendix~\ref{app:justification_single_step}.

Our DTD framework employs two distinct scoring branches to quantify anomalies based on the predicted noise \(\hat{\epsilon} = \epsilon_\theta(x_k, k, x_{hist})\), addressing different operational needs in UAV monitoring. The nonparametric scoring branch offers interpretable alternatives using statistical methods. 
The parametric scoring branch leverages an energy-based model for efficient and scalable anomaly detection, ideal for high-dimensional sensor data. Both approaches ensure robust detection by exploiting the diffusion model’s learned representation of normal data distributions.

\subsubsection{Nonparametric Scoring}

% The nonparametric scoring branch employs a versatile suite of statistical techniques to assign anomaly scores to the predicted noise \(\hat{\epsilon} = \epsilon_\theta(x_k, k, x_{hist})\), typically computed at a small time step \(k\) (e.g., \(k=1\)). For a test sample \(x\), we generate the perturbed sample \(x_k = \sqrt{\bar{\alpha}_k} x + \sqrt{1 - \bar{\alpha}_k} \epsilon\), with \(\epsilon \sim \mathcal{N}(0, I)\), and evaluate \(\hat{\epsilon} = \epsilon_\theta(x_k, k, x_{hist})\). These techniques assess deviations of \(\hat{\epsilon}\) from the normal data distribution learned by the diffusion model trained on \(p_{\text{data}}\). Normal data produce \(\hat{\epsilon}\) aligned with the expected distribution \(\mathcal{N}(0, I)\), resulting in low anomaly scores, while anomalous data, such as UAV sensor faults (e.g., engine power drops), yield diverging \(\hat{\epsilon}\), producing high scores. This approach ensures robust detection across diverse data modalities, including images, time series, and sensor readings.
The nonparametric branch scores the predicted noise \(\hat{\epsilon}=\epsilon_{\theta}(x_k,k,x_{\text{hist}})\), computed at a small step \(k\) (typically \(k=1\)). For a test sample \(x\) we form \(x_k=\sqrt{\bar{\alpha}_k}\,x+\sqrt{1-\bar{\alpha}_k}\,\epsilon\) with \(\epsilon\sim\mathcal{N}(0,I)\) and then compute \(\hat{\epsilon}\). Deviations of \(\hat{\epsilon}\) from the normal distribution of diffusion model yield high anomaly scores, while alignment with \(\mathcal{N}(0,I)\) yields low scores. This design provides robust detection across images, time series, and sensor streams.

The framework maintains a memory bank \(\{\hat{\epsilon}_i^+\}_{i=1}^M\) of predicted noises from normal training data, with \(\hat{\epsilon}_i^+=\epsilon_\theta(x_{k,i},k)\). We set \(k=0\) to capture the empirical distribution of normal noise predictions. Nonparametric scorers compare a test \(\hat{\epsilon}\) to this reference to quantify deviation, yielding low scores for normal data and high scores for anomalies across images and multivariate time series. Examples of applicable techniques include the following, tailored to experimental contexts:

\begin{itemize}
  \item \textbf{Kernel Density Estimation (KDE).} Gaussian kernel with bandwidth \(h=1.06\,\sigma M^{-1/5}\):
  \[
  s_{\text{KDE}}(\hat{\epsilon})=-\log\!\left(\frac{1}{M}\sum_{i=1}^M K_h(\hat{\epsilon},\hat{\epsilon}_i^+)+10^{-8}\right).
  \]
  Lower density implies a higher score.

  \item \textbf{k-Nearest Neighbors (kNN).} Mean distance to the \(k\) nearest neighbors:
  \[
  s_{\text{kNN}}(\hat{\epsilon})=\frac{1}{k}\sum_{j\in\mathcal{N}_k(\hat{\epsilon})}\|\hat{\epsilon}-\hat{\epsilon}_j^+\|_2.
  \]
  Larger distances imply a higher score.

  \item \textbf{Isolation Forest (iForest).} Inverse average path length:
  \[
  s_{\text{iForest}}(\hat{\epsilon})=2^{-\mathbb{E}[\text{path}(\hat{\epsilon})]/c(M)},\quad
  c(M)\approx 2\ln(M-1)+0.5772-\frac{2(M-1)}{M}.
  \]
  Shorter paths imply a higher score.
\end{itemize}

\subsubsection{Parametric EBM Scoring}

The parametric branch uses an Energy-Based Model (EBM) with parameters \(\phi\) to score the predicted noise \(\hat{\epsilon}=\epsilon_\theta(x_k,k,x_{\text{hist}})\) from a small diffusion step (typically \(k=1\)). For a test sample \(x\) with context \(x_{\text{hist}}\), we form \(x_k=\sqrt{\bar{\alpha}_k}\,x+\sqrt{1-\bar{\alpha}_k}\,\epsilon\) with \(\epsilon\sim\mathcal{N}(0,I)\). We then compute \(\hat{\epsilon}\) and evaluate the energy \(E_\phi(\hat{\epsilon})\). Lower energy indicates conformity to normal data. Higher energy flags anomalies.

\begin{proposition}
\label{prop:ebm_score}
Let the diffusion model be trained on \(p_{\text{data}}(x\mid x_{\text{hist}})\) with noise predictor \(\epsilon_\theta(x_k,k,x_{\text{hist}})\) and an EBM parameterized by \(\phi\). Then the energy \(E_\phi\!\big(\epsilon_\theta(x_k,k,x_{\text{hist}})\big)\) serves as a monotone proxy for the negative log-likelihood \textit{\(-\log p_{\theta,\phi}(x\mid x_{\text{hist}})\)} under the induced diffusion-based likelihood.
\end{proposition}

A derivation is provided in Appendix~\ref{app:ebm_score_derivation}. We construct \(p_{\theta,\phi}(x\mid x_{\text{hist}})\) by marginalizing over the noise and relate the EBM energy to \(-\log p_{\theta,\phi}\) up to an additive constant.
Intuitively, for normal UAV sensor data conditioned on historical context \(x_{hist}\), the predicted noise \(\hat{\epsilon} = \epsilon_\theta(x_k, k, x_{hist})\) aligns with the training distribution, yielding low \(E_\phi(\hat{\epsilon})\) due to the EBM optimization. Anomalous data, such as erratic sensor readings deviating from expected patterns given \(x_{hist}\), produce \(\hat{\epsilon}\) with high energy, enabling effective anomaly detection.

The EBM defines an unnormalized probability distribution over the predicted noise:

\begin{equation}
	p_\phi(\hat{\epsilon}) = \frac{1}{Z(\phi)} \exp\left[ f_\phi(\hat{\epsilon}) \right],
	\label{eq:ebm_dist}
\end{equation}
where \(f_\phi: \mathbb{R}^d \to \mathbb{R}\) is a negative energy function implemented as a multi-layer perceptron with parameters \(\phi\), and \(Z(\phi) = \int \exp\left[ f_\phi(\hat{\epsilon}) \right] d\hat{\epsilon}\) is the intractable normalizing constant. The anomaly score is defined as \(E_\phi(\hat{\epsilon}) = -f_\phi(\hat{\epsilon})\), where high \(E_\phi(\hat{\epsilon})\) (low \(f_\phi(\hat{\epsilon})\)) indicates anomalies, and low \(E_\phi(\hat{\epsilon})\) corresponds to normal data.

The EBM is trained contrastively to distinguish normal from anomalous noise predictions. Positive samples \(\hat{\epsilon}^+ = \epsilon_\theta(x_k, k)\) are generated from normal data \(x \sim p_{\text{data}}\). Negative samples \(\hat{\epsilon}^-\) are synthesized to be distinct from normal data using Markov chain Monte Carlo (MCMC) via Langevin dynamics, ensuring they lie further from the learned distribution. The dynamics iterate as:

\begin{equation}
	\hat{\epsilon}^-_m = \hat{\epsilon}^-_{m-1} + \frac{\delta^2}{2} \nabla_{\hat{\epsilon}} f_\phi(\hat{\epsilon}^-_{m-1}) + \delta \eta_m, \quad \eta_m \sim \mathcal{N}(0, I),
	\label{eq:langevin}
\end{equation}
where \(m\) indexes the MCMC step, \(\delta\) is the step size, \(\eta_m\) is Brownian noise, and the chain is initialized from a uniform distribution or perturbed normal samples. This process drives \(\hat{\epsilon}^-\) towards regions of lower probability under \(p_\phi\), enhancing their distinctiveness from \(\hat{\epsilon}^+\).
% The EBM loss is:

% \begin{equation}
% 	\mathcal{L}_{\text{EBM}} = \mathbb{E}_{x \sim p_{\text{data}}} \left[ E_\phi(\hat{\epsilon}^+) \right] - \mathbb{E}_{\hat{\epsilon}^- \sim p_{\text{neg}}} \left[ E_\phi(\hat{\epsilon}^-) \right],
% 	\label{eq:ebm_loss}
% \end{equation}

% where \(p_{\text{neg}}\) is the distribution of negative samples, approximated via Langevin dynamics. This minimizes energy for normal predictions while maximizing it for anomalies, ensuring robust scoring.

The parametric approach is computationally efficient, requiring only a single forward pass through the diffusion model and EBM, making it scalable for high-dimensional UAV sensor data. It excels in real-time applications, detecting faults like sudden power drops with low latency, enhancing safety in critical scenarios.

\subsection{Anomaly Labeling with EVT}\label{sec:anomaly_labeling}
We post-process DTD scores using Peaks Over Threshold (POT) from Extreme Value Theory. Scores \(s_t\) from the parametric and nonparametric branches are computed from the predicted noise \(\hat{\epsilon}=\epsilon_{\theta}(x_k,k,x_{\text{hist}})\) at a small step \(k\) (typically \(k=1\)) with \(x_k=\sqrt{\bar{\alpha}_k}\,x+\sqrt{1-\bar{\alpha}_k}\,\epsilon\). Using normal-training scores \(\{s_i^+\}_{i=1}^M\), we choose a high threshold \(t\), fit a Generalized Pareto Distribution to excesses \(s>t\), and set the adaptive decision level
\[
z_q = t + \frac{\hat{\sigma}}{\hat{\gamma}}\Bigl(\bigl(\tfrac{qM}{N_t}\bigr)^{-\hat{\gamma}} - 1\Bigr),
\]
where \(N_t\) is the number of excesses and \(q\) (\(10^{-3}\)) is a risk parameter. Test scores above \(z_q\) are labeled as anomalous. POT provides data-driven thresholds with small memory and low overhead, enabling reliable streaming and real-time UAV fault detection.

\section{Experimental Setup}

\subsection{Dataset Description}

To evaluate the DTD framework, we utilize four diverse datasets: the Air Lab Fault and Anomaly (ALFA) dataset, Biomisa Arducopter Sensory Critique (BASiC) dataset, the Server Machine Dataset (SMD), and the CIFAR-10 dataset. 
% These datasets span multiple data modalities (time series and images) and anomaly types, enabling a comprehensive assessment of DTD’s generalizability and robustness in detecting anomalies across varied domains.

The ALFA dataset, which is a real-world UAV corpus for fault and anomaly detection \citep{keipour2021} contains 47 autonomous flights with 66 minutes of normal data and 13 minutes of post-fault data. Scenarios cover 23 full engine failures and 24 control-surface faults (rudder, aileron, elevator). For ALFA, we built an automated preprocessing pipeline (to be released) that resamples to 50\,Hz, interpolates for temporal consistency, and drops the first 20\,s to remove early-flight instability such as takeoff and controller initialization, following \citealp{keipour2019}. Guided by domain work \citep{chen2024,jiang2024}, we select informative signals and organize them into 19 graph nodes that act as logical sensors, with 14 features per node to support spatiotemporal modeling. The processed data are serialized for reproducibility and split \textcolor{black}{70\%, 15\%, 15\%} into train, validation, and test. Fault timestamps and types provide ground truth for both DTD branches. To broaden evaluation beyond ALFA, we also use BASiC \citep{ahmad2024}, which offers 70 autonomous flights and more than seven hours of data.

SMD is a five-week collection of resource-utilization traces from 28 cluster machines. We evaluate two challenging sequences (Machine~1-1 and 2-1) highlighted by \cite{tuli2022}. The data are multivariate time series of system metrics such as CPU, memory, and disk I/O. Unlike ALFA’s sudden faults, SMD exhibits intermittent anomalies that can be sporadic or persistent, stressing temporal robustness. We split the dataset into 70\% training, 15\% validation, and 15\% test sets using the provided labels for evaluation.

% The CIFAR-10 dataset, a standard benchmark for image classification, consists of 60,000 32x32 color images across 10 classes (e.g., airplanes, cars)~\citep{krizhevsky2009}. To adapt CIFAR-10 for anomaly detection, we employ a one-vs-rest strategy, treating one class as normal and others as anomalous. Images are preprocessed using ResNet-152 to extract high-level features, reducing dimensionality while preserving semantic information, and then fed into DTD for anomaly scoring. The dataset is split into 70\% training, 15\% validation, and 15\% testing, with class labels repurposed to define normal and anomalous instances. This setup tests DTD’s ability to handle image-based anomalies, complementing the time-series focus of ALFA, BASiC, and SMD.
CIFAR-10 contains 60{,}000 \(32\times32\) color images across 10 classes \citep{krizhevsky2009}. We use a one-vs-rest setup with one class as normal and the rest as anomalous. Images are embedded with ResNet-152 to obtain high-level features, then scored by DTD. The split is 70\% train, 15\% validation, and 15\% test, with labels repurposed to define normal vs.\ anomalous instances. This setting evaluates DTD on image anomalies and complements the time-series datasets.

% We selected these datasets for two primary reasons. \textcolor{black}{First, their diverse modalities, time series (ALFA, BASiC, SMD) and images (CIFAR-10), demonstrate DTD’s generalizability,} because UAV systems may encounter similar data formats (e.g., sensor streams, onboard imagery). Second, they present varied anomaly challenges: ALFA captures sudden, critical UAV faults (e.g., engine failures), while SMD includes intermittent anomalies within normal operations, and CIFAR-10 tests static image-based anomalies. This diversity ensures DTD is robust not only for UAV-specific sudden faults but also for complex, non-persistent anomalies across domains, validating its applicability in real-world scenarios with heterogeneous data. For detailed statistics of the raw data, refer to Appendix~\ref{app:dataset_stats}.
We selected these datasets for two reasons. First, their modality diversity, with time series (ALFA, BASiC, SMD) and images (CIFAR 10), shows DTD’s generalizability to UAV relevant formats such as sensor streams and onboard image. Second, they present different anomaly profiles, where ALFA captures sudden faults, SMD contains intermittent anomalies during normal operation, and CIFAR 10 tests static image anomalies. This diversity shows robustness across heterogeneous data and real world scenarios. Detailed statistics are in Appendix~\ref{app:dataset_stats}.

\subsection{Evaluation Metrics}
We evaluate the DTD framework using standard anomaly detection metrics: precision, recall, F1-score, and accuracy. Details for these metrics are provided in Appendix~\ref{app:evaluation_metrics}.
% We evaluate the DTD framework using standard anomaly detection metrics: precision, recall, F1-score, and accuracy, defined as:
The experiments were conducted on an NVIDIA A6000 workstation equipped with 48 GB of GPU memory and 128 GB of RAM. To ensure robustness, all experiments were replicated five times with different random seeds, and the results were averaged. The code and datasets will be made publicly available upon publication.

\section{UAV Case Study: Results and Discussion}

% We evaluate DTD’s parametric (DM-P) and nonparametric (DM-NP) branches on four datasets to test generalizability across modalities and anomaly types, including UAV faults, time-series anomalies, and image outliers. Results for each modality are presented below, followed by a summary discussion.

We first evaluate DTD’s parametric (DM-P) and nonparametric (DM-NP) branches on UAV sensor data, while later sections report results on other modalities.

% We evaluate DTD’s parametric (DM-P) and nonparametric (DM-NP) branches on four datasets 
% % : ALFA and BASiC (UAV sensor data, graph-structured), Server Machine Dataset (SMD, multivariate time series), and CIFAR-10 (images). 
% to test DTD’s generalizability across diverse modalities and anomaly types, from abrupt UAV faults to intermittent time-series anomalies and static image outliers. 
% The following subsections detail the results for each modality, followed by a discussion synthesizing the findings.

\subsection{Performance Metrics and Visualization of Anomaly Scores}

The left column of Table~\ref{tab:alfa_basic_results_combined} presents a comparative evaluation of the DTD framework’s branches against established baselines, including the ALFA baseline~\citep{keipour2021} and an advanced Transformer-based approach~\citep{ahmad2024}. The results underscore the exceptional performance of DTD’s branches, with DM-NP demonstrating leading accuracy and robustness, closely followed by DM-P, both surpassing the baselines in effectively identifying anomalies while minimizing false positives and negatives. As shown in Table~\ref{tab:alfa_basic_results_combined}, we observe similar trends in the BASiC dataset, where DM-NP and DM-P outperform the Transformer baseline, particularly in precision and F1-score. Though the Transformer baseline exhibits superior recall on this dataset, this advantage comes at the expense of significantly lower precision and F1-score, whereas DM-NP and DM-P achieve a more balanced and reliable performance across all metrics, reinforcing their suitability for robust UAV anomaly detection across diverse datasets.

\begin{table}[t]
	\centering
	\footnotesize
	\caption{Performance on ALFA and BASIC datasets (UAV sensor data, graph-structured). 
	Reported values are the mean $\pm$ standard deviation over 5 runs.}	
	\label{tab:alfa_basic_results_combined}
	\resizebox{\textwidth}{!}{
	\begin{tabular}{@{}lcccccc@{}}
	\toprule
	\textbf{Method} & \multicolumn{3}{c}{\textbf{ALFA}} & \multicolumn{3}{c}{\textbf{BASIC}} \\
	\cmidrule(lr){2-4} \cmidrule(lr){5-7}
	 & \textbf{Prec.} & \textbf{Rec.} & \textbf{F1} & \textbf{Prec.} & \textbf{Rec.} & \textbf{F1} \\
	\midrule
	DM-NP & \textbf{0.9914}$\pm$0.006 & \textbf{0.9890}$\pm$0.006 & \textbf{0.9901}$\pm$0.005 & \textbf{0.9897}$\pm$0.014 & 0.9415$\pm$0.018 & \textbf{0.9648}$\pm$0.009 \\
	DM-P & \textbf{0.9943}$\pm$0.008 & \textbf{0.9678}$\pm$0.006 & \textbf{0.9807}$\pm$0.004 & \textbf{0.9911}$\pm$0.015 & 0.9027$\pm$0.032 & \textbf{0.9443}$\pm$0.018 \\
	ALFA Base.~\citep{keipour2021} & 0.7365 & 0.9115 & 0.8119 & -- & -- & -- \\
	Transformer~\citep{ahmad2024} & 0.8720 & 0.9437 & 0.8928 & 0.5360 & \textbf{1.0000} & 0.6979 \\
	\bottomrule
	\end{tabular}
	}
	\end{table}

\begin{figure*}
	\centering
	\begin{minipage}[t]{0.48\textwidth}
		\centering
		\captionsetup[subfloat]{font=tiny,labelfont=tiny}
		\subfloat[DM-NP (KDE) Scores (ALFA)]{
			\includegraphics[width=0.99\linewidth]{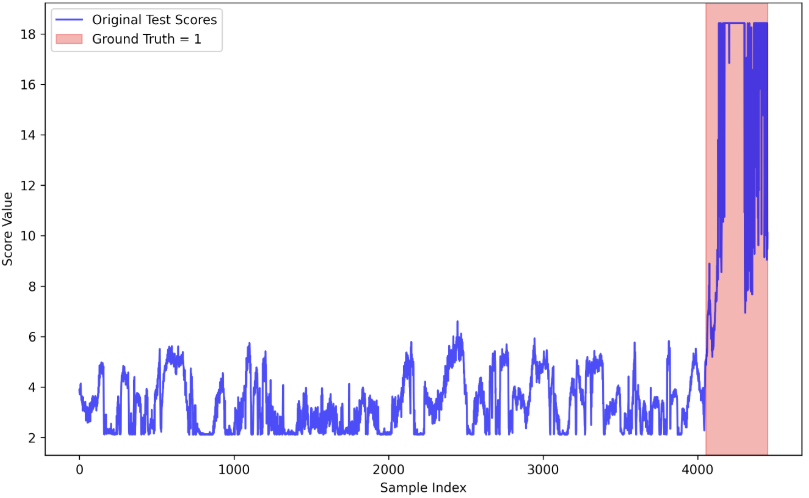}
			\label{fig:dmkde_scores_alfa}
		}
		\hfill
		\subfloat[DM-P Scores (ALFA)]{
			\includegraphics[width=0.99\linewidth]{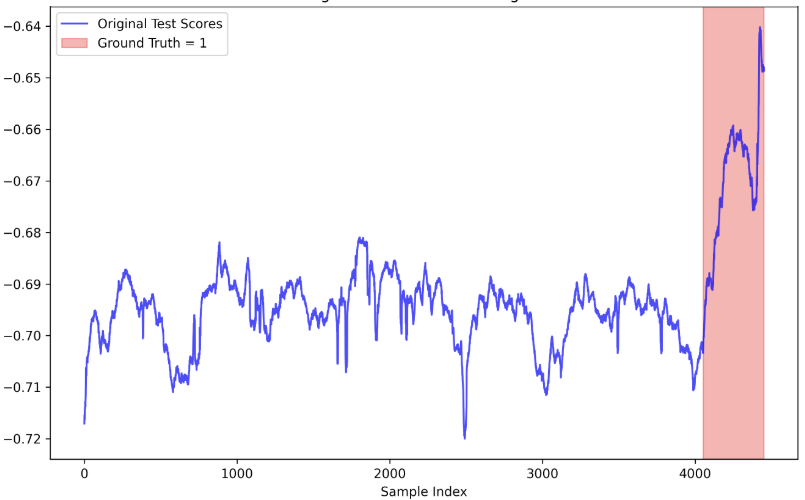}
			\label{fig:dmebm_scores_alfa}
		}
	\end{minipage}
	\hfill
	\begin{minipage}[t]{0.48\textwidth}
		\centering
		\captionsetup[subfloat]{font=tiny,labelfont=tiny}

		\subfloat[DM-NP (KDE) Predictions (ALFA)]{
			\includegraphics[width=0.99\linewidth]{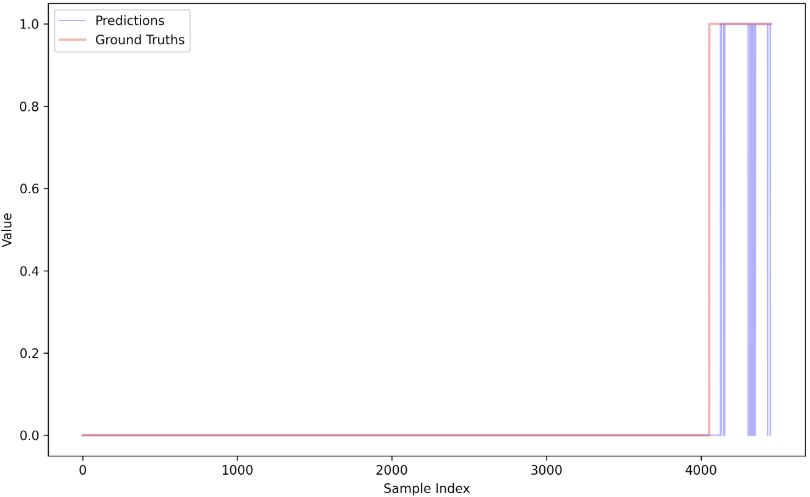}
			\label{fig:dmkde_predictions_alfa}
		}
		\hfill
		\subfloat[DM-EBM Predictions (ALFA)]{
			\includegraphics[width=0.99\linewidth]{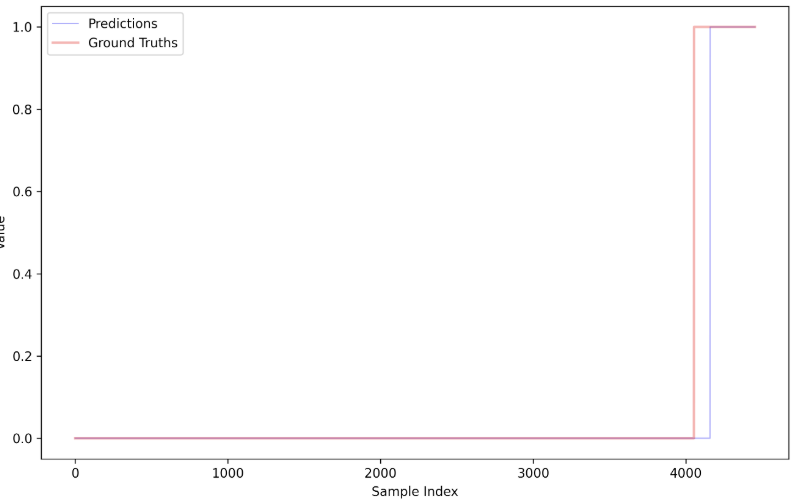}
			\label{fig:dmebm_predictions_alfa}
		}
	\end{minipage}
	\caption{Anomaly scores and predictions for ALFA dataset. Left column (a–b): scores; right column (c-d): predictions. Ground truth (fault) regions are shaded in red.}
	\label{fig:combined_results}
\end{figure*}

	% Figure~\ref{fig:combined_results} illustrates the anomaly scores and predictions generated by the DTD framework for the ALFA dataset, highlighting the effectiveness of both nonparametric (DM-NP) and parametric (DM-P) scoring branches in detecting anomalies. 
	For the ALFA dataset (Flight 0), where an engine failure occurs around sample index 4000, Figure~\ref{fig:combined_results} (a) and Figure~\ref{fig:combined_results} (b) depict the DM-NP and DM-P anomaly scores, respectively, with score increases aligning well with the ground truth regions shaded in red. Figure~\ref{fig:combined_results} (c) and Figure~\ref{fig:combined_results} (d) show the corresponding predictions, where both branches accurately capture the fault onset. The DTD framework achieves zero false positives in the ALFA dataset, a critical advantage for safety-critical UAV applications. While some scores in fault regions may appear less pronounced, the framework prioritizes detection accuracy over score magnitude. As long as scores exceed the threshold set by Section~\ref{sec:anomaly_labeling}, anomalies are reliably identified, meeting the task’s requirements, though minor detection delays may occur in rare cases. Similar performance is observed in the BASiC dataset, visualized in Appendix~\ref{app:additional_visualizations}
.

\subsection{Analysis of Node Connection Changes in UAV Sensor Data}

% \begin{figure*}[t]
% 	\centering
% 	\includegraphics[width=0.99\textwidth]{figures/DMKDE_UAV_nodes_connection.png}
% 	\caption{Visualization of UAV nodes connection in the DM-P (KDE) for graph-structured ALFA sensor data.}
% 	\label{fig:uav_nodes_connection}
% 	figures/DMKDE_UAV_nodes_connection_a_Pre-Transition Changes (Period 2 - Period 1).png
% 	figures/b_Critical Transition Changes(Period 3 - Period 2).png
% 	figures/c_Post-Transition Changes (Period 4 - Period 3).png
% \end{figure*}

\begin{figure*}[t]
	\centering
	\begin{minipage}[t]{\textwidth}
		\centering
		\captionsetup[subfloat]{font=tiny,labelfont=tiny} 
		\subfloat[Pre-Transition Changes]{
			\includegraphics[width=0.32\linewidth]{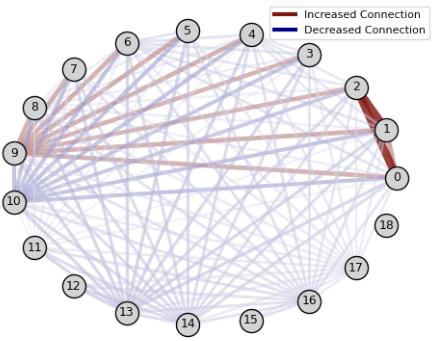}
			\label{subfig:pre_transition}
		}
		\hfill
		\subfloat[Critical Transition Changes]{
			\includegraphics[width=0.32\linewidth]{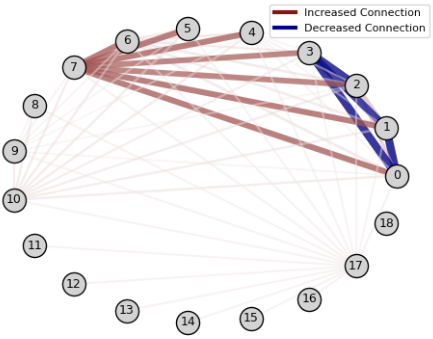}
			\label{subfig:critical_transition}
		}
		\hfill
		\subfloat[Post-Transition Changes]{
			\includegraphics[width=0.32\linewidth]{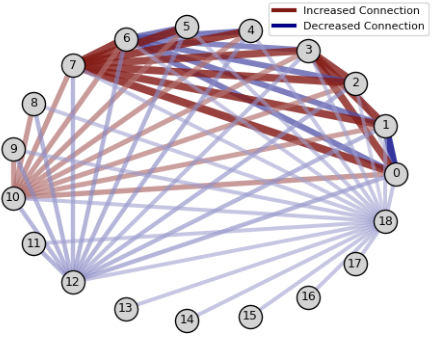}
			\label{subfig:post_transition}
		}
	\end{minipage}
	\caption{Visualization of UAV nodes connection in the DM-P (KDE) for graph-structured ALFA sensor data.}
	\label{fig:uav_nodes_connection}
\end{figure*}

Using ALFA’s graph-structured UAV data, we analyze how DTD tracks inter-sensor relationships across three phases (Pre-Transition, Critical-Transition, Post-Transition). As shown in Figure~\ref{fig:uav_nodes_connection}, nodes are sensors (Table~\ref{tab:sensors}, Appendix~\ref{app:sensor_descriptions}); red/blue edges mark strengthened/weakened connections between successive phases. Changes are minimal pre-transition, localized at onset, and widespread post-transition, reflecting the anomaly’s expanding impact on sensor dynamics.

% This analysis explores how the DTD framework captures evolving relationships between UAV sensors using graph-structured data from the ALFA dataset across three phases: Pre-Transition, Critical Transition, and Post-Transition. In Figure~\ref{fig:uav_nodes_connection}, nodes represent sensors (see Table~\ref{tab:sensors} in Appendix~\ref{app:sensor_descriptions} for full descriptions of sensors), with red edges indicating increased connections and blue edges showing decreased connections between phases. Connection changes are least pronounced in the Pre-Transition phase, become more noticeable for a few nodes during the Critical Transition, and are most pronounced in the Post-Transition phase, reflecting the evolving impact of the anomaly on sensor dynamics.

\subsubsection{Pre-Transition Changes}
During stable flight, connectivity shifts are small. Figure~\ref{fig:uav_nodes_connection} (a) shows slight strengthening from node~2 (local position) to nodes~1 (flight dynamics) and~0 (navigation errors), consistent with routine stabilization against minor drifts. Modest gains around nodes~8 (global position) and~9 (magnetic field) likely reflect benign navigation adjustments.
% During stable flight, adjustments are minor. The Pre-Transition Figure~\ref{fig:uav_nodes_connection}a highlights slight increases in connections between node 2 (local position, e.g., velocity, orientation) and nodes 1 (flight dynamics, e.g., altitude, speed) and 0 (navigation errors), as indicated by red edges. This increasing connectivity suggests the UAV’s navigation system integrates local position data with flight dynamics and navigation error corrections to maintain stability, likely in response to minor environmental factors such as positional drifts. Other connections, such as those around nodes 8 (global position data) and 9 (magnetic field data), show minor increases, possibly due to navigation adjustments for magnetic interference.

\subsubsection{Critical Transition Changes}
At anomaly onset (e.g., engine failure), Figure~\ref{fig:uav_nodes_connection} (b) shows strong connection increases between node~7 (control outputs) and nodes~4--6 (wind, GPS velocity, measured velocity), indicating rapid fusion to counter thrust/stability loss. Meanwhile, node~3 (IMU) shows weakend connections toward nodes~0--2 (navigation errors, flight dynamics, local position), suggesting disrupted IMU reliability and a shift to other cues.
% This phase captures the onset of an anomaly, engine failure in the example. The Critical Transition Figure~\ref{fig:uav_nodes_connection}b shows pronounced increases in connections for node 7 (control outputs, e.g., throttle), with red edges connecting to nodes 4 (wind estimation), 5 (GPS velocity), and 6 (measured velocity). This reflects the control system’s rapid integration of environmental and velocity data to mitigate thrust and stability loss. Conversely, node 3 (IMU data, e.g., acceleration) exhibits decreased connections with nodes 0 (navigation errors), 1 (flight dynamics), and 2 (local position), as indicated by blue edges. This reduction suggests disrupted dependencies, likely because IMU data becomes less reliable or relevant during the anomaly, prompting the system to prioritize other inputs.

\subsubsection{Post-Transition Changes}
As recovery begins, connectivity reaches peak. Figure~\ref{fig:uav_nodes_connection} (c) shows node~7 (control outputs) strengthening broadly to nodes~0--6 (navigation errors, flight dynamics, local position, IMU, wind, GPS velocity, measured velocity), indicating system-wide fusion to regain stability. By contrast, node~12 (yaw) weakens toward nodes~0--2, consistent with prioritizing velocity/position over heading.
% As the UAV stabilizes, changes become most pronounced. The Post-Transition Figure~\ref{fig:uav_nodes_connection}c reveals significant increases in connections for node 7 (control outputs), with red edges connecting to nodes 0 (navigation errors), 1 (flight dynamics), 2 (local position), 3 (IMU data), 4 (wind estimation), 5 (GPS velocity), and 6 (measured velocity). This widespread integration underscores the control system’s reliance on a broad range of sensor data to restore stability. Meanwhile, node 12 (yaw angles) shows decreased connections with nodes 0 (navigation errors), 1 (flight dynamics), and 2 (local position), as indicated by blue edges, reflecting reduced interdependence as the UAV prioritizes velocity and position corrections over heading adjustments.
% \textcolor{black}{The trend of least pronounced changes before transition, more noticeable changes for a few nodes during transition, and most pronounced changes after transition highlights the DTD framework’s ability to track sensor dynamics during anomalies.}

\subsubsection{Summary of Sensor Relationship Dynamics}
Sensor relationships shift with operating phase. 
\textcolor{black}{Changes are minimal at the begining, accurately localized when anomaly occurs, and broadest after anomaly happens, highlighting DTD’s tracking capability of anomaly-driven sensor dynamics.}
% For example, connections between flight dynamics (node~1) and local position (node~2) strengthen near anomalies, while navigation velocity (node~6) and control outputs (node~7) intensify during recovery. 
\textcolor{black}{Such non-static dependencies motivate our adaptive, data-driven graph learning (Section~\ref{sec:methodology}). 
\textcolor{black}{
Rather than fixing the graph, DTD learns sensor relationships via GNNs and captures both spatial (inter-sensor) and temporal dependencies, thereby improving sensitivity to subtle faults and providing a comprehensive view of the UAV’s state.}
% DTD infers relations rather than fixing the graph and, with GNNs, captures both spatial (inter-sensor) and temporal dependencies, boosting sensitivity to subtle faults and providing a holistic view of the UAV’s state.
}
% \textcolor{black}{The observed dynamics in sensor relationships, such as increased connections between flight dynamics (node 1) and local position data (node 2) during anomalies, and between navigation velocity (node 6) and control outputs (node 7) during recovery, show that these dependencies vary with operational conditions. This variability underscores their non-static nature.} This variability reinforces the necessity of our data-driven, adaptive approach, which infers sensor relationships dynamically rather than relying on a fixed graph structure, as discussed in Section~\ref{sec:methodology}. By leveraging GNNs within this framework, DTD effectively captures both spatial (inter-sensor) and temporal dependencies, enhancing its sensitivity to subtle anomalies and providing a comprehensive view of the UAV's operational state.

% \begin{figure*}[t]
% 	\centering
% 	\includegraphics[width=0.8\textwidth]{figures/DMKDE_UAV_nodes_heatmap.png}
% 	\caption{Visualization of UAV nodes heatmap in the DMKDE framework for graph-structured sensor data.}
% 	\label{fig:uav_nodes_connection}
% \end{figure*}

\subsection{Comparative Energy Assignment Analysis of DM-NP and DM-P Branches}

\begin{figure*}[ht]
	\centering
	\begin{minipage}[t]{0.92\textwidth}
		\centering
		\captionsetup[subfloat]{font=tiny,labelfont=tiny} % Set font size for both caption and label
		\subfloat[DM-NP (KDE) Surface Plot]{
			\includegraphics[width=0.45\linewidth]{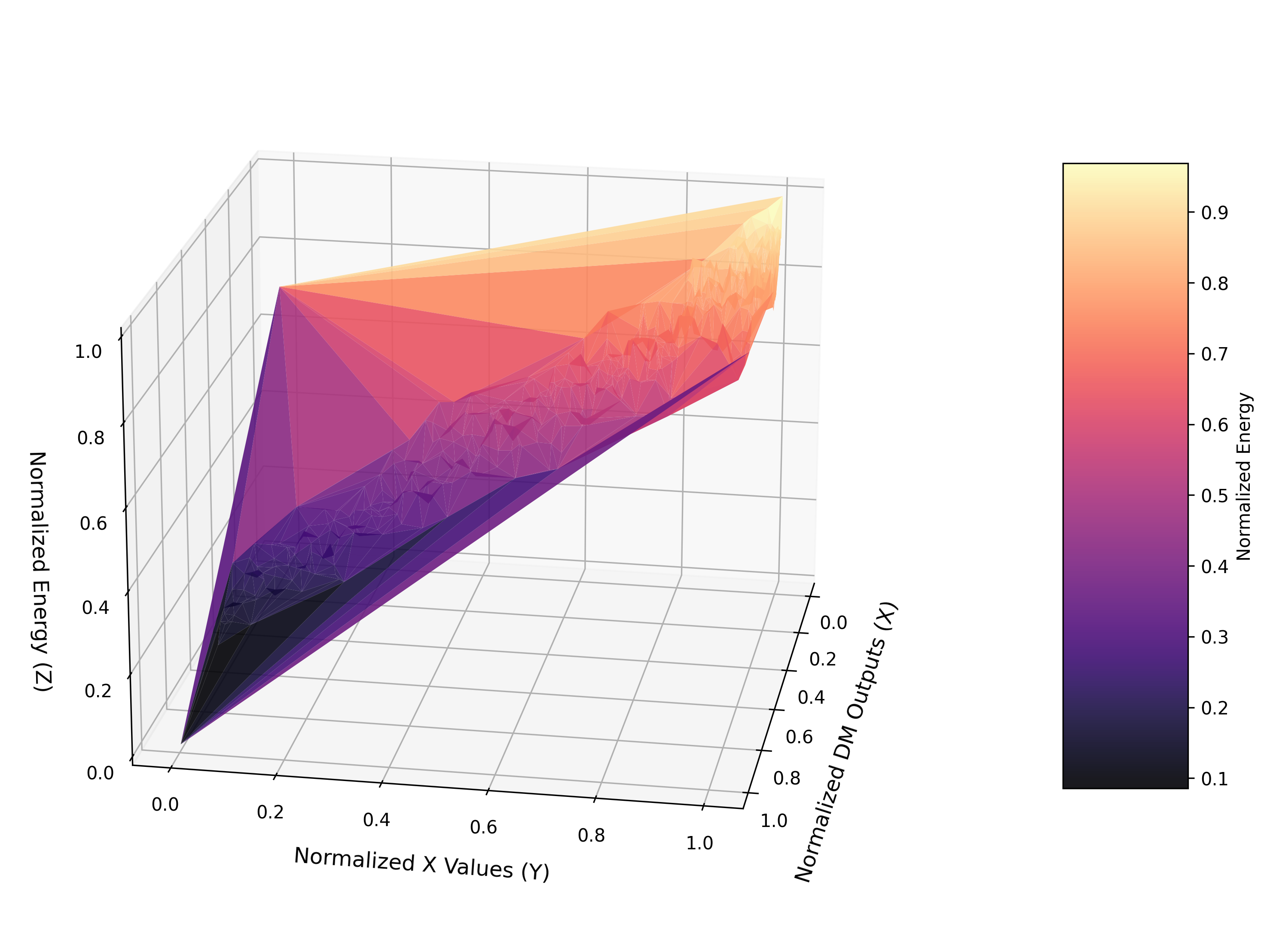}
			\label{fig:dmkde_3d_surface}
		}
		\hfill
		\subfloat[DM-P Surface Plot]{
			\includegraphics[width=0.45\linewidth]{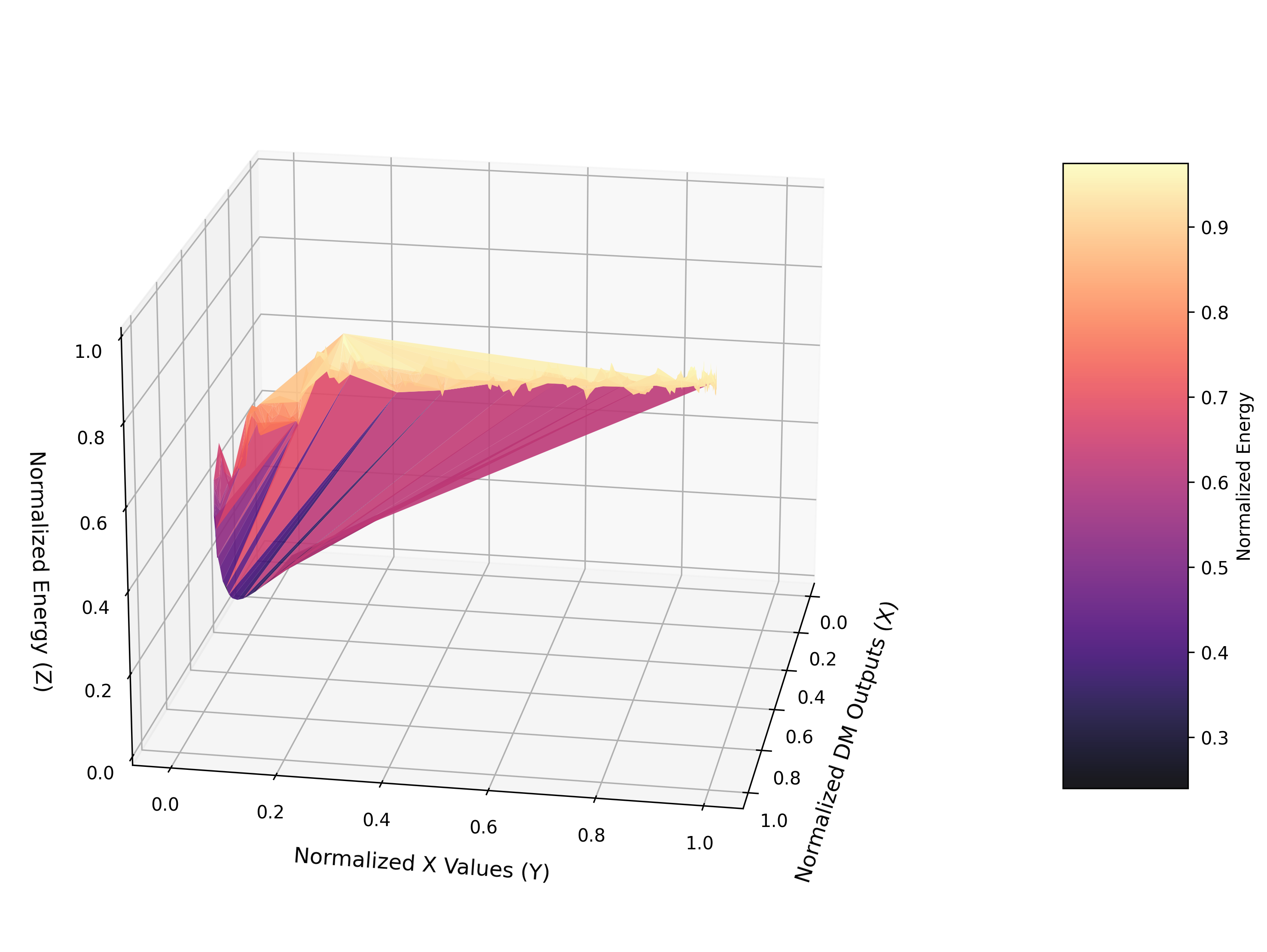}
			\label{fig:dmebm_3d_surface}
		}
	\end{minipage}
	\caption{3D surface plots of DM outputs, diffused data and normalized energy values for both branches on ALFA dataset: (a) DM-NP (KDE) and (b) DM-P. The \emph{x}-axis represents noise prediction outputs, the \emph{y}-axis represents diffused values (0 to 1), and the \emph{z}-axis represents energy outputs.}
	\label{fig:3d_surface_visualizations}
\end{figure*}

Figure~\ref{fig:3d_surface_visualizations} compares DM-NP (KDE) and DM-P via 3D surfaces with diffusion level on the \(x\)-axis, noise-prediction output on the \(y\)-axis, and energy on the \(z\)-axis (higher implies more likely anomalous). DM-NP shows a sharp high-energy peak at large diffusion with elevated noise outputs, while DM-P rises more gradually and spreads energy across a wider range. 
Notably, for DM-NP, when no noise is added (\emph{y} = 0), some diffusion outputs are non-zero. By contrast, DM-P produces near zero diffusion outputs in this case, but DM-NP’s assigned energy remains near zero, correctly identifying normal data.
This highlights that DM outputs alone are insufficient for anomaly detection, emphasizing the necessity of the scoring mechanism. Both branches effectively assign high energy to diffused anomalous data, with DM-NP's concentrated peak and DM-P's wider distribution showcasing their complementary strengths.

% Figure~\ref{fig:3d_surface_visualizations} presents two 3D surface plots comparing the DM-NP (KDE) and DM-P branches, with diffused values (\emph{y}-axis, from 0 for normal to maximum diffusion), energy model (\emph{z}-axis, higher values indicating anomaly likelihood), and DM noise prediction outputs (\emph{y}-axis). The DM-NP (KDE) plot shows a sharp peak at higher diffused values (y near 1.0) and energy (z near 1.0), with elevated noise outputs, indicating strong, localized high-energy assignment to anomalous data, whereas the DM-P plot exhibits an elongated surface with a gradual rise, distributing high energy more broadly across diffused values. 
% % {Notably, for DM-NP, when no noise is added (Y = 0), some DM outputs are non-zero as expected, unlike DM-P where outputs are zero, potentially a drawback for DM-NP; however, DM-NP's assigned energy remains near zero, correctly identifying normal data. }
% Notably, for DM-NP, when no noise is added (y = 0), some diffusion outputs are non-zero. By contrast, DM-P produces near zero diffusion outputs in this case, but DM-NP’s assigned energy remains near zero, correctly identifying normal data.
% This highlights that DM outputs alone are insufficient for anomaly detection, emphasizing the necessity of the scoring mechanism. Both branches effectively assign high energy to diffused anomalous data, with DM-NP's concentrated peak and DM-P's wider distribution showcasing their complementary strengths.

\section{Evaluation on Additional Modalities and More Visualizations}
To assess the versatility of the DTD framework beyond UAV sensor data, we evaluate its performance on two distinct modalities: multivariate time series data from SMD and image data from the CIFAR-10 dataset.
We also provide additional plots in Appendix~\ref{app:gradient_dynamics} that visualizes DM-P (EBM) on SMD, covering the PCA of the raw data, one-step diffusion, the energy gradient at \(k=1\), and example trajectories of normal and anomalous points. These plots collectively demonstrate our framework’s ability to distinguish normal from anomalous data.

\subsection{Evaluation on Multivariate Time Series Data}

Unlike ALFA/BASiC, which we model as graphs, SMD is evaluated as raw multivariate time series, demonstrating DTD’s adaptability without graph construction. Against recent time-series baselines TranAD~\citep{tuli2022} and D3R~\citep{wang2023e}, DTD achieves near-perfect metrics and consistently outperforms them (as shown in Table~\ref{tab:smd_results_combined}). TranAD sometimes excels in precision but at the cost of recall and F1-Score. Further visualizations of anomaly scores and predictions are provided in Appendix~\ref{app:additional_visualizations}.

\begin{table}[t]
	\centering
	\footnotesize
	\caption{Performance on SMD dataset (multivariate time series). Reported values are the mean $\pm$ standard deviation over 5 independent runs.}
	\label{tab:smd_results_combined}
	\begin{tabular}{@{}lcccccc@{}}
	\toprule
	\textbf{Method} & \multicolumn{3}{c}{\textbf{Machine 1-1}} & \multicolumn{3}{c}{\textbf{Machine 2-1}} \\
	\cmidrule(lr){2-4} \cmidrule(lr){5-7}
	 & \textbf{Prec.} & \textbf{Rec.} & \textbf{F1} & \textbf{Prec.} & \textbf{Rec.} & \textbf{F1} \\
	\midrule
	DM-NP (KDE) & \textbf{0.9947}$\pm$0.0034 & \textbf{0.9993}$\pm$0.0011 & \textbf{0.9970}$\pm$0.0016 & 0.9821$\pm$0.0056 & \textbf{0.9945}$\pm$0.0031 & \textbf{0.9883}$\pm$0.0025 \\
	DM-NP (KNN) & \textbf{0.9975}$\pm$0.0025 & \textbf{0.9979}$\pm$0.0012 & \textbf{0.9977}$\pm$0.0008 & 0.9702$\pm$0.0123 & \textbf{0.9932}$\pm$0.0000 & \textbf{0.9815}$\pm$0.0063 \\
	DM-P & \textbf{0.9965}$\pm$0.0018 & \textbf{0.9996}$\pm$0.0007 & \textbf{0.9980}$\pm$0.0008 & 0.9815$\pm$0.0064 & \textbf{0.9925}$\pm$0.0045 & \textbf{0.9869}$\pm$0.0027 \\
	TranAD~\citep{tuli2022} & 0.9026 & 0.9974 & 0.9476 & \textbf{1.0000} & 0.3744 & 0.5448 \\
	D3R~\citep{wang2023e} & 0.8589 & 0.9935 & 0.9213 & 0.7482 & 0.9790 & 0.8482 \\
	\bottomrule
	\end{tabular}
	\end{table}

\subsection{Evaluation on Image Data}

% \begin{table}[t]
% 	\centering
% 	\caption{Performance on CIFAR-10 dataset (images).}
% 	\label{tab:cifar10_results}
% 	\begin{tabular}{@{}l@{\hspace{5pt}}c@{\hspace{5pt}}c@{\hspace{5pt}}c@{}}
% 		\toprule
% 		\textbf{Method}          & \textbf{Precision}                      & \textbf{Accuracy}                & \textbf{F1-score}                       \\
% 		\midrule
% 		DM-NP (iForest)         & \textbf{0.9948}{\scriptsize$\pm$0.0004} & {0.9593}{\scriptsize$\pm$0.0026} & \textbf{0.9693}{\scriptsize$\pm$0.0017} \\
% 		DM-P            & \textbf{0.9942}{\scriptsize$\pm$0.0002} & {0.9547}{\scriptsize$\pm$0.0013} & \textbf{0.9665}{\scriptsize$\pm$0.0012} \\
% 		NeuTraL-F~\citep{qiu2025} & --                                      & {0.9530}{\scriptsize$\pm$0.0600} & --                                      \\
% 		PANDA~\citep{reiss2021a}  & --                                      & \textbf{0.9620}                  & --                                      \\
% 		\bottomrule
% 	\end{tabular}
% \end{table}

\begin{table}[t]
	\centering
	\footnotesize
	\caption{Performance on CIFAR-10 dataset (images). Reported values are the mean $\pm$ standard deviation over 5 independent runs.}
	\label{tab:cifar10_results}
	\begin{tabular}{@{}lccc@{}}
	\toprule
	\textbf{Method} & \textbf{Prec.} & \textbf{Acc.} & \textbf{F1} \\
	\midrule
	DM-NP (iForest) & \textbf{0.9948}$\pm$0.0004 & 0.9593$\pm$0.0026 & \textbf{0.9693}$\pm$0.0017 \\
	DM-P & \textbf{0.9942}$\pm$0.0002 & 0.9547$\pm$0.0013 & \textbf{0.9665}$\pm$0.0012 \\
	NeuTraL-F~\citep{qiu2025} & -- & 0.9530$\pm$0.0600 & -- \\
	PANDA~\citep{reiss2021a} & -- & \textbf{0.9620} & -- \\
	\bottomrule
	\end{tabular}
	\end{table}

The CIFAR-10 dataset \citep{krizhevsky2009} evaluates DTD for image anomaly detection in a one-vs-rest setup. Table~\ref{tab:cifar10_results} compares our parametric (DM-P) and nonparametric (DM-NP) branches with NeuTraL-F \citep{qiu2025} and PANDA \citep{reiss2021a}. DTD attains the best overall performance, with DM-NP yielding the top F1-Score and precision, and DM-P closely matching PANDA while surpassing NeuTraL-F. Unlike PANDA, which requires computationally intensive kNN-based detection at test time and storage of the full training dataset, DTD leverages efficient single-pass diffusion model evaluations, enabling rapid inference suitable for real-time applications, as described in Section~\ref{sec:methodology}~\citep{reiss2021a}. 
These results complement our structured-data findings and underscore DTD’s cross-modal effectiveness.

\section{Conclusion}
% Anomaly detection is critical for ensuring safety and reliability in complex systems, particularly in applications like UAVs and industrial monitoring.
% The proposed DTD framework provides an effective approach to anomaly detection, 
% utilizing diffusion models to identify anomalies across diverse data types, including UAV sensor data, time series, and images. Its dual-branch design, with parametric (DM-P) and nonparametric (DM-NP) scoring methods, provides flexibility in addressing computational efficiency and interpretability. Our approach is supported by comprehensive theoretical underpinnings, which validate the use of predicted noise as a proxy for anomaly scoring. 
% The testing results
% on ALFA, BASiC, SMD, and CIFAR-10 show superior performance compared to existing methods, highlighting its robustness and generalization capabilities. The framework well addresses challenges in terms of high-dimensional data and complex dependencies, making it suitable for real-world applications. 
% Future work may extend this foundational approach to more intricate data structures, enhancing wide utility in safety-critical domains.
Anomaly detection is critical for safety and reliability in complex systems such as UAVs and industrial monitoring. We introduced DTD, a diffusion based framework that detects anomalies across UAV sensor data, multivariate time series, and images. The framework offers two scoring branches: parametric (DM-P) for efficiency and nonparametric (DM-NP) for interpretability, and our theory justifies using predicted noise as a principled anomaly signal. Across ALFA, BASiC, SMD, and CIFAR 10, DTD outperforms strong baselines and remains robust to high dimensional data and complex dependencies. DTD is practical for real systems, and future work will extend it to richer data structures and broader safety critical deployments.

%%%%%%%%%%%%%%%%%%%%%%%%%%%%%%%%%%%%%%%%%%%%%%%%%%%%%%%% APPENDIX %%%%%%%%%%%%%%%%%%%%%%%%%%%%%%%%%%%%%%%%%%%%%%%%%%%%%%%%

% % Acknowledgments here
% \ACKNOWLEDGMENT{We would like to express our sincere gratitude to [acknowledge individuals, organizations, or institutions] for their invaluable contributions to this research. We are also grateful to [mention any additional acknowledgements, such as technical assistance, data providers, or colleagues] for their support and assistance throughout the course of this work.}

% References here (outcomment the appropriate case)

% CASE 1: BiBTeX used to constantly update the references
%   (while the paper is being written).
\bibliographystyle{informs2014} % outcomment this and next line in Case 1
\bibliography{references} % if more than one, comma separated

\begin{thebibliography}{37}
\providecommand{\natexlab}[1]{#1}
\providecommand{\url}[1]{\texttt{#1}}
\providecommand{\urlprefix}{URL }

\bibitem[{Ahmad et~al.(2024)Ahmad, Akram, Mohsan, Saghar, Ahmad,
  \protect\BIBand{} Butt}]{ahmad2024}
Ahmad MW, Akram MU, Mohsan MM, Saghar K, Ahmad R, Butt WH (2024)
  Transformer-based sensor failure prediction and classification framework for
  {{UAVs}}. \emph{Expert Systems with Applications} 248:123415, ISSN 09574174,
  \urlprefix\url{http://dx.doi.org/10.1016/j.eswa.2024.123415}.

\bibitem[{Bai et~al.(2020)Bai, Yao, Li, Wang, \protect\BIBand{} Wang}]{bai2020}
Bai L, Yao L, Li C, Wang X, Wang C (2020) Adaptive {{Graph Convolutional
  Recurrent Network}} for {{Traffic Forecasting}}.
  \urlprefix\url{http://arxiv.org/abs/2007.02842}.

\bibitem[{{Ben-Gal} et~al.(2023){Ben-Gal}, Bacher, Amara, \protect\BIBand{}
  Shmueli}]{ben-gal2023}
{Ben-Gal} I, Bacher M, Amara M, Shmueli E (2023) A {{Nonparametric Subspace
  Analysis Approach}} with {{Application}} to {{Anomaly Detection Ensembles}}.
  \emph{INFORMS Journal on Data Science} 2(2):99--115, ISSN 2694-4022,
  2694-4030, \urlprefix\url{http://dx.doi.org/10.1287/ijds.2023.0027}.

\bibitem[{Chen et~al.(2024)Chen, Lyu, Shi, \protect\BIBand{} Zhang}]{chen2024}
Chen H, Lyu Y, Shi J, Zhang W (2024) {{UAV Anomaly Detection Method Based}} on
  {{Convolutional Autoencoder}} and {{Support Vector Data Description}} with
  0/1 {{Soft-Margin Loss}}. \emph{Drones} 8(10):534, ISSN 2504-446X,
  \urlprefix\url{http://dx.doi.org/10.3390/drones8100534}.

\bibitem[{Deng et~al.(2024)Deng, Lu, Tao, Liu, \protect\BIBand{}
  Chen}]{deng2024}
Deng H, Lu Y, Tao Y, Liu Z, Chen J (2024) Unmanned {{Aerial Vehicles}} anomaly
  detection model based on sensor information fusion and hybrid multimodal
  neural network. \emph{Engineering Applications of Artificial Intelligence}
  132:107961, \urlprefix\url{http://dx.doi.org/10.1016/j.engappai.2024.107961}.

\bibitem[{Dhakal et~al.(2023)Dhakal, Bosma, Chaudhary, \protect\BIBand{}
  Kandel}]{dhakal2023a}
Dhakal R, Bosma C, Chaudhary P, Kandel LN (2023) {{UAV Fault}} and {{Anomaly
  Detection Using Autoencoders}}. \emph{2023 {{IEEE}}/{{AIAA}} 42nd {{Digital
  Avionics Systems Conference}} ({{DASC}})}, 1--8, ISSN 2155-7209,
  \urlprefix\url{http://dx.doi.org/10.1109/DASC58513.2023.10311126}.

\bibitem[{Efron(2011)}]{efron2011}
Efron B (2011) Tweedie's {{Formula}} and {{Selection Bias}}. \emph{Journal of
  the American Statistical Association} 106(496):1602--1614, ISSN 0162-1459,
  \urlprefix\url{https://www.ncbi.nlm.nih.gov/pmc/articles/PMC3325056/}.

\bibitem[{He et~al.(2022)He, Wei, \protect\BIBand{} Wen}]{he2022a}
He M, Wei Z, Wen JR (2022) Convolutional {{Neural Networks}} on {{Graphs}} with
  {{Chebyshev Approximation}}, {{Revisited}}. \emph{The Thirty-Sixth Annual
  Conference on Neural Information Processing Systems} .

\bibitem[{Ho et~al.(2020)Ho, Jain, \protect\BIBand{} Abbeel}]{ho2020}
Ho J, Jain A, Abbeel P (2020) Denoising {{Diffusion Probabilistic Models}}.
  \emph{{{NeurIPS}}} (arXiv), \urlprefix\url{http://arxiv.org/abs/2006.11239}.

\bibitem[{Ho et~al.(2025)Ho, Karami, \protect\BIBand{} Armanfard}]{ho2025}
Ho TKK, Karami A, Armanfard N (2025) Graph {{Anomaly Detection}} in {{Time
  Series}}: {{A Survey}}.
  \urlprefix\url{http://dx.doi.org/10.48550/arXiv.2302.00058}.

\bibitem[{Hu et~al.(2020)Hu, Gao, Li, Wu, Du, \protect\BIBand{}
  Maybank}]{hu2020}
Hu W, Gao J, Li B, Wu O, Du J, Maybank S (2020) Anomaly {{Detection Using Local
  Kernel Density Estimation}} and {{Context-Based Regression}}. \emph{IEEE
  Transactions on Knowledge and Data Engineering} 32(2):218--233, ISSN
  1041-4347, 1558-2191, 2326-3865,
  \urlprefix\url{http://dx.doi.org/10.1109/TKDE.2018.2882404}.

\bibitem[{Huang et~al.(2025)Huang, Wang, Pei, Wang, Alexanian,
  \protect\BIBand{} Niyato}]{huang2025}
Huang H, Wang P, Pei J, Wang J, Alexanian S, Niyato D (2025) Deep {{Learning
  Advancements}} in {{Anomaly Detection}}: {{A Comprehensive Survey}}.
  \urlprefix\url{http://dx.doi.org/10.48550/arXiv.2503.13195}.

\bibitem[{Huet et~al.(2022)Huet, Navarro, \protect\BIBand{} Rossi}]{huet2022}
Huet A, Navarro JM, Rossi D (2022) Local {{Evaluation}} of {{Time Series
  Anomaly Detection Algorithms}}. \emph{Proceedings of the 28th {{ACM SIGKDD
  Conference}} on {{Knowledge Discovery}} and {{Data Mining}}}, 635--645,
  \urlprefix\url{http://dx.doi.org/10.1145/3534678.3539339}.

\bibitem[{Jiang et~al.(2024)Jiang, Nan, Zhang, Li, \protect\BIBand{}
  Li}]{jiang2024}
Jiang G, Nan P, Zhang J, Li Y, Li X (2024) Robust {{Spatial-Temporal
  Autoencoder}} for {{Unsupervised Anomaly Detection}} of {{Unmanned Aerial
  Vehicle With Flight Data}}. \emph{IEEE Transactions on Instrumentation and
  Measurement} 73:1--14, ISSN 0018-9456, 1557-9662,
  \urlprefix\url{http://dx.doi.org/10.1109/TIM.2024.3428649}.

\bibitem[{Keipour et~al.(2019)Keipour, Mousaei, \protect\BIBand{}
  Scherer}]{keipour2019}
Keipour A, Mousaei M, Scherer S (2019) Automatic {{Real-time Anomaly
  Detection}} for {{Autonomous Aerial Vehicles}}. \emph{2019 {{International
  Conference}} on {{Robotics}} and {{Automation}} ({{ICRA}})}, 5679--5685,
  \urlprefix\url{http://dx.doi.org/10.1109/ICRA.2019.8794286}.

\bibitem[{Keipour et~al.(2021)Keipour, Mousaei, \protect\BIBand{}
  Scherer}]{keipour2021}
Keipour A, Mousaei M, Scherer S (2021) {{ALFA}}: {{A}} dataset for {{UAV}}
  fault and anomaly detection. \emph{The International Journal of Robotics
  Research} 40(2-3):515--520, ISSN 0278-3649, 1741-3176,
  \urlprefix\url{http://dx.doi.org/10.1177/0278364920966642}.

\bibitem[{Kim \protect\BIBand{} Kim(2023)}]{kim2023a}
Kim H, Kim H (2023) Contextual anomaly detection for high-dimensional data
  using {{Dirichlet}} process variational autoencoder. \emph{IISE Transactions}
  55(5):433--444, ISSN 2472-5854, 2472-5862,
  \urlprefix\url{http://dx.doi.org/10.1080/24725854.2021.2024925}.

\bibitem[{Krizhevsky(2009)}]{krizhevsky2009}
Krizhevsky A (2009) Learning {{Multiple Layers}} of {{Features}} from {{Tiny
  Images}} .

\bibitem[{Lin et~al.(2024)Lin, Liu, Wu, \protect\BIBand{} Zhao}]{lin2024}
Lin H, Liu G, Wu J, Zhao JL (2024) Deterring the {{Gray Market}}: {{Product
  Diversion Detection}} via {{Learning Disentangled Representations}} of
  {{Multivariate Time Series}}. \emph{INFORMS Journal on Computing}
  36(2):571--586, ISSN 1091-9856,
  \urlprefix\url{http://dx.doi.org/10.1287/ijoc.2022.0155}.

\bibitem[{Lin et~al.(2010)Lin, Khalastchi, \protect\BIBand{} Kaminka}]{lin2010}
Lin R, Khalastchi E, Kaminka GA (2010) Detecting anomalies in unmanned vehicles
  using the {{Mahalanobis}} distance. \emph{2010 {{IEEE International
  Conference}} on {{Robotics}} and {{Automation}}}, 3038--3044 (Anchorage, AK:
  IEEE), ISBN 978-1-4244-5038-1,
  \urlprefix\url{http://dx.doi.org/10.1109/ROBOT.2010.5509781}.

\bibitem[{Liu et~al.(2023)Liu, Zhu, Feng, \protect\BIBand{} Zeng}]{liu2023g}
Liu F, Zhu X, Feng P, Zeng L (2023) Anomaly {{Detection}} via {{Progressive
  Reconstruction}} and {{Hierarchical Feature Fusion}}. \emph{Sensors}
  23(21):8750, ISSN 1424-8220,
  \urlprefix\url{http://dx.doi.org/10.3390/s23218750}.

\bibitem[{Livernoche et~al.(2023)Livernoche, Jain, Hezaveh, \protect\BIBand{}
  Ravanbakhsh}]{livernoche2023}
Livernoche V, Jain V, Hezaveh Y, Ravanbakhsh S (2023) On {{Diffusion Modeling}}
  for {{Anomaly Detection}}. \emph{{{ICLR}} 2024}, 31,
  \urlprefix\url{https://openreview.net/forum?id=lR3rk7ysXz}.

\bibitem[{Ma et~al.(2023)Ma, Wu, Xue, Yang, Zhou, Sheng, Xiong,
  \protect\BIBand{} Akoglu}]{ma2023a}
Ma X, Wu J, Xue S, Yang J, Zhou C, Sheng QZ, Xiong H, Akoglu L (2023) A
  {{Comprehensive Survey}} on {{Graph Anomaly Detection}} with {{Deep
  Learning}}. \emph{IEEE Transactions on Knowledge and Data Engineering}
  35(12):12012--12038, ISSN 1041-4347, 1558-2191, 2326-3865,
  \urlprefix\url{http://dx.doi.org/10.1109/TKDE.2021.3118815}.

\bibitem[{Qiu et~al.(2025)Qiu, Kloft, Mandt, \protect\BIBand{}
  Rudolph}]{qiu2025}
Qiu C, Kloft M, Mandt S, Rudolph M (2025) Self-{{Supervised Anomaly Detection
  With Neural Transformations}}. \emph{IEEE Transactions on Pattern Analysis
  and Machine Intelligence} 47(3):2170--2185, ISSN 1939-3539,
  \urlprefix\url{http://dx.doi.org/10.1109/TPAMI.2024.3519543}.

\bibitem[{Reiss et~al.(2021)Reiss, Cohen, Bergman, \protect\BIBand{}
  Hoshen}]{reiss2021a}
Reiss T, Cohen N, Bergman L, Hoshen Y (2021) {{PANDA}}: {{Adapting Pretrained
  Features}} for {{Anomaly Detection}} and {{Segmentation}}. \emph{2021
  {{IEEE}}/{{CVF Conference}} on {{Computer Vision}} and {{Pattern
  Recognition}} ({{CVPR}})}, 2805--2813 (Nashville, TN, USA: IEEE), ISBN
  978-1-6654-4509-2,
  \urlprefix\url{http://dx.doi.org/10.1109/CVPR46437.2021.00283}.

\bibitem[{Shakhatreh et~al.(2019)Shakhatreh, Sawalmeh, {Al-Fuqaha}, Dou,
  Almaita, Khalil, Othman, Khreishah, \protect\BIBand{}
  Guizani}]{shakhatreh2019}
Shakhatreh H, Sawalmeh AH, {Al-Fuqaha} A, Dou Z, Almaita E, Khalil I, Othman
  NS, Khreishah A, Guizani M (2019) {Unmanned Aerial Vehicles (UAVs): A Survey
  on Civil Applications and Key Research Challenges}. \emph{IEEE Access}
  7:48572--48634, ISSN 2169-3536,
  \urlprefix\url{http://dx.doi.org/10.1109/ACCESS.2019.2909530}.

\bibitem[{Suboh et~al.(2023)Suboh, Aziz, Shaharudin, Ismail, \protect\BIBand{}
  Mahdin}]{suboh2023}
Suboh S, Aziz IA, Shaharudin SM, Ismail SA, Mahdin H (2023) A {{Systematic
  Review}} of {{Anomaly Detection}} within {{High Dimensional}} and
  {{Multivariate Data}}. \emph{JOIV : International Journal on Informatics
  Visualization} 7(1):122, ISSN 2549-9904, 2549-9610,
  \urlprefix\url{http://dx.doi.org/10.30630/joiv.7.1.1297}.

\bibitem[{Tao \protect\BIBand{} Du(2025)}]{tao2025}
Tao C, Du J (2025) {{PointSGRADE}}: {{Sparse}} learning with graph
  representation for anomaly detection by using unstructured {{3D}} point cloud
  data. \emph{IISE Transactions} 57(2):131--144, ISSN 2472-5854, 2472-5862,
  \urlprefix\url{http://dx.doi.org/10.1080/24725854.2023.2285840}.

\bibitem[{Tuli et~al.(2022)Tuli, Casale, \protect\BIBand{} Jennings}]{tuli2022}
Tuli S, Casale G, Jennings NR (2022) {{TranAD}}: {{Deep Transformer Networks}}
  for {{Anomaly Detection}} in {{Multivariate Time Series Data}}.
  \emph{{{VLDB}} 2022} (arXiv),
  \urlprefix\url{http://dx.doi.org/10.48550/arXiv.2201.07284}.

\bibitem[{Vaswani et~al.(2017)Vaswani, Shazeer, Parmar, Uszkoreit, Jones,
  Gomez, Kaiser, \protect\BIBand{} Polosukhin}]{NIPS2017_3f5ee243}
Vaswani A, Shazeer N, Parmar N, Uszkoreit J, Jones L, Gomez AN, Kaiser Lu,
  Polosukhin I (2017) Attention is all you need. Guyon I, Luxburg UV, Bengio S,
  Wallach H, Fergus R, Vishwanathan S, Garnett R, eds., \emph{Advances in
  Neural Information Processing Systems}, volume~30 (Curran Associates, Inc.),
  \urlprefix\url{https://proceedings.neurips.cc/paper_files/paper/2017/file/3f5ee243547dee91fbd053c1c4a845aa-Paper.pdf}.

\bibitem[{Wang et~al.(2019)Wang, Liu, Peng, \protect\BIBand{} Wang}]{wang2019b}
Wang B, Liu D, Peng X, Wang Z (2019) Data-{{Driven Anomaly Detection}} of
  {{UAV}} based on {{Multimodal Regression Model}}. \emph{2019 {{IEEE
  International Instrumentation}} and {{Measurement Technology Conference}}
  ({{I2MTC}})}, 1--6, ISSN 2642-2077,
  \urlprefix\url{http://dx.doi.org/10.1109/I2MTC.2019.8827154}.

\bibitem[{Wang et~al.(2023)Wang, Zhuang, Qi, Wang, Wang, Sun, \protect\BIBand{}
  Liao}]{wang2023e}
Wang C, Zhuang Z, Qi Q, Wang J, Wang X, Sun H, Liao J (2023) Drift doesn't
  {{Matter}}: {{Dynamic Decomposition}} with {{Diffusion Reconstruction}} for
  {{Unstable Multivariate Time Series Anomaly Detection}}. \emph{{{NeurIPS}}
  2023}, 17.

\bibitem[{Xu et~al.(2025)Xu, Wang, Zhou, \protect\BIBand{} Lu}]{xu2025}
Xu J, Wang J, Zhou Z, Lu T (2025) Toward {{Graph Data Collaboration}} in a
  {{Data-Sharing-Free Manner}}: {{A Novel Privacy-Preserving Graph Pretraining
  Model}}. \emph{INFORMS Journal on Computing} ijoc.2023.0115, ISSN 1091-9856,
  1526-5528, \urlprefix\url{http://dx.doi.org/10.1287/ijoc.2023.0115}.

\bibitem[{Yang et~al.(2024)Yang, Li, Zhang, Zhu, \protect\BIBand{}
  Liao}]{yang2024e}
Yang L, Li S, Zhang Y, Zhu C, Liao Z (2024) Deep {{Learning-Assisted Unmanned
  Aerial Vehicle Flight Data Anomaly Detection}}: {{A Review}}. \emph{IEEE
  Sensors Journal} 24(20):31681--31695, ISSN 1558-1748,
  \urlprefix\url{http://dx.doi.org/10.1109/JSEN.2024.3451648}.

\bibitem[{Yu et~al.(2020)Yu, Lv, Tong, \protect\BIBand{} Dong}]{yu2020}
Yu Y, Lv P, Tong X, Dong J (2020) Anomaly {{Detection}} in {{High-Dimensional
  Data Based}} on {{Autoregressive Flow}}. Nah Y, Cui B, Lee SW, Yu JX, Moon
  YS, Whang SE, eds., \emph{Database {{Systems}} for {{Advanced
  Applications}}}, volume 12113, 125--140 (Cham: Springer International
  Publishing), ISBN 978-3-030-59415-2 978-3-030-59416-9,
  \urlprefix\url{http://dx.doi.org/10.1007/978-3-030-59416-9_8}.

\bibitem[{Zhang(2024)}]{zhang2024b}
Zhang H (2024) An attempt to generate new bridge types from latent space of
  generative flow. \urlprefix\url{http://dx.doi.org/10.48550/arXiv.2401.10299},
  arXiv preprint.

\bibitem[{Zhou et~al.(2024)Zhou, He, Chen, \protect\BIBand{} Chang}]{zhou2024a}
Zhou S, He Z, Chen X, Chang W (2024) An {{Anomaly Detection Method}} for {{UAV
  Based}} on {{Wavelet Decomposition}} and {{Stacked Denoising Autoencoder}}.
  \emph{Aerospace} 11(5):393, ISSN 2226-4310,
  \urlprefix\url{http://dx.doi.org/10.3390/aerospace11050393}.

\end{thebibliography}

%\bibliographystyle{informs2014} % outcomment this and next line in Case 1
%\bibliography{sample} % if more than one, comma separated

% CASE 2: BiBTeX used to generate mypaper.bbl (to be further fine tuned)
%\input{mypaper.bbl} % outcomment this line in Case 2

%If you don't use BiBTex, you can manually itemize references as shown below.

%\bibliographystyle{nonumber}

% \begin{thebibliography}{3}
% \providecommand{\natexlab}[1]{#1}
% \providecommand{\url}[1]{\texttt{#1}}
% \providecommand{\urlprefix}{URL }

% \bibitem[{Smith(2005)}]{smith2005}
% Smith J (2005) Optimal resource allocation in humanitarian logistics.
%   \emph{Journal of Operations Research} 30(2):123--135.
  
% \bibitem[{Jones(2010)}]{jones2010}
% Jones S (2010) Stochastic programming models for humanitarian logistics.
%   \emph{INFORMS Mathematics of Operations Research} 35(4):567--580.

% \bibitem[{Brown(2015)}]{brown2015}
% Brown D (2015) \emph{Introduction to Stochastic Programming} (Springer).

% \end{thebibliography}

% %\THEEndNotes
% \begingroup \parindent 0pt \parskip 0.0ex \def\enotesize{\normalsize} \theendnotes \endgroup

% Appendix here
% Options are (1) APPENDIX (with or without general title) or
%             (2) APPENDICES (if it has more than one unrelated sections)
% Outcomment the appropriate case if necessary
%
% \begin{APPENDIX}{<Title of the Appendix>}
% \end{APPENDIX}
%
%   or
%
\clearpage
\begin{APPENDICES}
    \section{Derivation of Noise Prediction as Scaled Score Function}
    \label{app:proof_noise_score}
    
    This appendix provides the detailed proof of Proposition~\ref{prop:noise_score}, which states that the predicted noise \(\epsilon_\theta(x_k, k, x_{hist})\) in a diffusion model, conditioned on historical data \(x_{hist}\), approximates a scaled score function, \(\epsilon_\theta(x_k, k, x_{hist}) \approx -\sqrt{1 - \bar{\alpha}_k} \nabla_{x_k} \log p_k(x_k | x_{hist})\), and at small \(k\), \(\epsilon_\theta(x_0, 0, x_{hist}) \approx -\sqrt{1 - \bar{\alpha}_0} \nabla_{x_0} \log p_{\text{data}}(x_0 | x_{hist})\). The proof leverages the forward process of the diffusion model, Tweedie’s formula, and the training objective to establish this relationship, justifying the use of predicted noise for anomaly detection in contexts where historical data informs the model.
    
    We begin with the forward process defined in Equation~\eqref{eq:forward_diffusion} as follows:
    \begin{equation*}
        x_k = \sqrt{\bar{\alpha}_k} x_0 + \sqrt{1 - \bar{\alpha}_k} \epsilon, \quad \epsilon \sim \mathcal{N}(0, I),
    \end{equation*}
    where \(x_0 \sim p_{\text{data}}(\cdot | x_{hist})\), \(\bar{\alpha}_k = \prod_{s=0}^k (1 - \beta_s)\), and \(\beta_s \in (0, 1)\) is the noise schedule. The data distribution \(p_{\text{data}}(x_0 | x_{hist})\) is conditioned on historical data \(x_{hist}\), reflecting the temporal context of the system, such as past UAV sensor readings. The neural network \(\epsilon_\theta(x_k, k, x_{hist})\) is trained to predict \(\epsilon\), minimizing the objective in Equation~\eqref{eq:diffusion_loss} as follows:
    
    \begin{equation*}
        \mathcal{L}_{\text{DM}} = \mathbb{E}_{x_0 \sim p_{\text{data}}(\cdot | x_{hist}), \epsilon \sim \mathcal{N}(0, I), k} \left[ \|\epsilon - \epsilon_\theta(x_k, k, x_{hist})\|_2^2 \right].
    \end{equation*}
    
    To show that \(\epsilon_\theta(x_k, k, x_{hist})\) approximates a scaled score, we first derive the conditional expectation of the noise given \(x_k\) and \(x_{hist}\). Rearranging Equation~\eqref{eq:forward_diffusion} for \(x_0\):
    
    \begin{equation}
        x_0 = \frac{x_k - \sqrt{1 - \bar{\alpha}_k} \epsilon}{\sqrt{\bar{\alpha}_k}}.
        \label{eq:app_x0_rearrange}
    \end{equation}
    
    Taking the conditional expectation given \textcolor{black}{the fixed} \(x_k\) and \(x_{hist}\):
    
    \begin{align}
        \mathbb{E}[x_0 | x_k, x_{hist}] &= \mathbb{E} \left[ \frac{x_k - \sqrt{1 - \bar{\alpha}_k} \epsilon}{\sqrt{\bar{\alpha}_k}} \Bigg| x_k, x_{hist} \right] \notag \\
        &= \frac{x_k}{\sqrt{\bar{\alpha}_k}} - \frac{\sqrt{1 - \bar{\alpha}_k}}{\sqrt{\bar{\alpha}_k}} \mathbb{E}[\epsilon | x_k, x_{hist}],
        \label{eq:app_cond_expect_x0}
    \end{align}
    
    Solving for \(\mathbb{E}[\epsilon | x_k, x_{hist}]\):
    
    \begin{equation}
        \mathbb{E}[\epsilon | x_k, x_{hist}] = \frac{x_k - \sqrt{\bar{\alpha}_k} \mathbb{E}[x_0 | x_k, x_{hist}]}{\sqrt{1 - \bar{\alpha}_k}}.
        \label{eq:app_expected_noise}
    \end{equation}
    
    The optimal noise predictor, minimizing the expected squared error, is \(\epsilon_\theta^*(x_k, k, x_{hist}) = \mathbb{E}[\epsilon | x_k, x_{hist}]\). Thus, the training objective ensures \(\epsilon_\theta(x_k, k, x_{hist}) \approx \epsilon_\theta^*(x_k, k, x_{hist})\) for normal data conditioned on \(x_{hist}\).
    
    Next, we connect \(\mathbb{E}[\epsilon | x_k, x_{hist}]\) to the score function \(\nabla_{x_k} \log p_k(x_k | x_{hist})\) using Tweedie’s formula~\citep{ho2020}. For a Gaussian perturbation \(x = z + \sigma \epsilon\), Tweedie’s formula states:
    
    \begin{equation}
        \nabla_x \log p(x) = \frac{1}{\sigma^2} \left( \mathbb{E}[z | x] - x \right).
        \label{eq:tweedie}
    \end{equation}
    
    In our model, \(x_k = \sqrt{\bar{\alpha}_k} x_0 + \sqrt{1 - \bar{\alpha}_k} \epsilon\), where \(z = \sqrt{\bar{\alpha}_k} x_0\) and \(\sigma = \sqrt{1 - \bar{\alpha}_k}\). Conditioning on \(x_{hist}\), we apply Tweedie’s formula to the conditional distribution \(p_k(x_k | x_{hist})\):
    
    \begin{equation}
        \nabla_{x_k} \log p_k(x_k | x_{hist}) = \frac{1}{1 - \bar{\alpha}_k} \left( \mathbb{E}[\sqrt{\bar{\alpha}_k} x_0 | x_k, x_{hist}] - x_k \right).
        \label{eq:app_tweedie_applied}
    \end{equation}
    
    \textcolor{black}{As} \(\mathbb{E}[\sqrt{\bar{\alpha}_k} x_0 | x_k, x_{hist}] = \sqrt{\bar{\alpha}_k} \mathbb{E}[x_0 | x_k, x_{hist}]\), we have:
    
    \begin{equation}
        \nabla_{x_k} \log p_k(x_k | x_{hist}) = \frac{1}{1 - \bar{\alpha}_k} \left( \sqrt{\bar{\alpha}_k} \mathbb{E}[x_0 | x_k, x_{hist}] - x_k \right).
        \label{eq:app_score_function}
    \end{equation}
    
    From Equation~\eqref{eq:app_expected_noise}, compute:
    
    \begin{align}
        x_k - \sqrt{\bar{\alpha}_k} \mathbb{E}[x_0 | x_k, x_{hist}] &= \sqrt{1 - \bar{\alpha}_k} \mathbb{E}[\epsilon | x_k, x_{hist}] \notag \\
        &= \sqrt{1 - \bar{\alpha}_k} \epsilon_\theta^*(x_k, k, x_{hist}).
        \label{eq:app_diff_expect}
    \end{align}
    
    Substitute into the score function:
    
    \begin{align}
        \nabla_{x_k} \log p_k(x_k | x_{hist}) &= \frac{1}{1 - \bar{\alpha}_k} \left( -\sqrt{1 - \bar{\alpha}_k} \epsilon_\theta^*(x_k, k, x_{hist}) \right) \notag \\
        &= -\frac{\epsilon_\theta^*(x_k, k, x_{hist})}{\sqrt{1 - \bar{\alpha}_k}}.
        \label{eq:app_score_substituted}
    \end{align}
    
    Thus:
    
    \begin{equation}
        \epsilon_\theta^*(x_k, k, x_{hist}) = -\sqrt{1 - \bar{\alpha}_k} \nabla_{x_k} \log p_k(x_k | x_{hist}).
        \label{eq:app_optimal_noise}
    \end{equation}
    
    Since \(\epsilon_\theta(x_k, k, x_{hist}) \approx \epsilon_\theta^*(x_k, k, x_{hist})\) due to training, we obtain:
    
    \begin{equation}
        \epsilon_\theta(x_k, k, x_{hist}) \approx -\sqrt{1 - \bar{\alpha}_k} \nabla_{x_k} \log p_k(x_k | x_{hist}).
        \label{eq:app_noise_score}
    \end{equation}
    
    For small \(k\), such as \(k=1\), \(\bar{\alpha}_0 \approx 1\), so \(x_0 \approx x_k\), and \(p_k(x_k | x_{hist}) \approx p_{\text{data}}(x_0 | x_{hist})\). Hence:
    
    \begin{equation}
        \epsilon_\theta(x_0, 0, x_{hist}) \approx -\sqrt{1 - \bar{\alpha}_0} \nabla_{x_0} \log p_{\text{data}}(x_0 | x_{hist}).
        \label{eq:app_small_k}
    \end{equation}
    
    This completes the first part of Proposition~\ref{prop:noise_score}, showing that \(\epsilon_\theta(x_k, k, x_{hist})\) encodes the geometry of the conditional data distribution \(p_k(x_k | x_{hist})\) via the score function.
    
    To validate consistency, we examine the training objective in Equation~\eqref{eq:diffusion_loss}. Substituting \(\epsilon = \frac{x_k - \sqrt{\bar{\alpha}_k} x_0}{\sqrt{1 - \bar{\alpha}_k}}\) from Equation~\eqref{eq:forward_diffusion} into the loss:
    
    \begin{equation}
        \small
        \mathcal{L}_{\text{DM}} = \mathbb{E}_{x_0 \sim p_{\text{data}}(\cdot | x_{hist}), \epsilon \sim \mathcal{N}(0, I), k} \left[ \left\| \frac{x_k - \sqrt{\bar{\alpha}_k} x_0}{\sqrt{1 - \bar{\alpha}_k}} - \epsilon_\theta(x_k, k, x_{hist}) \right\|_2^2 \right].
        \label{eq:app_loss_substituted}
    \end{equation}
    
    Taking expectations over \(\epsilon\), the optimal \(\epsilon_\theta(x_k, k, x_{hist})\) matches \(\mathbb{E}[\epsilon | x_k, x_{hist}]\). 
	% From Equations~\eqref{eq:app_expected_noise} and \eqref{eq:app_noise_score}, \(\mathbb{E}[\epsilon | x_k, x_{hist}] = -\sqrt{1 - \bar{\alpha}_k} \nabla_{x_k} \log p_k(x_k | x_{hist})\), confirming that the training aligns \(\epsilon_\theta(x_k, k, x_{hist})\) with the scaled negative score conditioned on \(x_{hist}\).
	From Equations~\eqref{eq:app_expected_noise} and \eqref{eq:app_noise_score}, the expected noise $\mathbb{E}[\epsilon | x_k, x_{hist}] = -\sqrt{1 - \bar{\alpha}_k} \nabla_{x_k} \log p_k(x_k | x_{hist})$ confirms that the training aligns $\epsilon_\theta(x_k, k, x_{hist})$ with the scaled negative score conditioned on $x_{hist}$. 
	This resembles denoising score matching, where the conditional score \(\nabla_{x_k} \log p_k(x_k | x_{hist})\) is approximated, reinforcing the theoretical foundation.
    
    Thus, Proposition~\ref{prop:noise_score} holds, as \(\epsilon_\theta(x_k, k, x_{hist})\) reliably approximates the scaled score of the conditional distribution, enabling anomaly detection by identifying deviations from the learned data distribution given historical context.

    \section{Derivation of EBM Scoring Likelihood}
    \label{app:ebm_score_derivation}
	\textcolor{black}{This appendix derives the diffusion-based likelihood used in Proposition~\ref{prop:ebm_score} and shows that the EBM energy \(E_\phi(\hat{\epsilon})\) approximates the negative log-likelihood \(-\log p_{\theta,\phi}(x \mid x_{\text{hist}})\) for anomaly detection. Here \(\hat{\epsilon}=\epsilon_\theta(x_k,k,x_{\text{hist}})\).}
    % This appendix derives the diffusion-based likelihood for Proposition~\ref{prop:ebm_score}, which states that the EBM energy score \(E_\phi(\hat{\epsilon})\), where \(\hat{\epsilon} = \epsilon_\theta(x_k, k, x_{hist})\), approximates the negative log-likelihood \(-\log p_{\theta, \phi}(x | x_{hist})\) for anomaly detection. 
	The derivation integrates the diffusion model’s noise prediction, conditioned on historical data \(x_{hist}\), with the EBM’s energy function, providing a probabilistic justification for the parametric scoring mechanism.
    
    We define the diffusion-based likelihood as a marginal over the noise \(\epsilon\):
    
    \begin{equation}
        p_{\theta, \phi}(x | x_{hist}) = \int p(\epsilon) p_{\theta, \phi}(x | \epsilon, x_{hist}) \, d\epsilon,
        \label{eq:cond_marginal_likelihood}
    \end{equation}
    where \(p(\epsilon) = \mathcal{N}(\epsilon | 0, I)\) is the noise prior, and \(p_{\theta, \phi}(x | \epsilon, x_{hist})\) is the conditional likelihood. For a test sample \(x\), the noisy sample at time step \(k\) is based on the forward diffusion process in Equation~\eqref{eq:forward_diffusion} as follows:
    
    \begin{equation*}
        x_k = \sqrt{\bar{\alpha}_k} x + \sqrt{1 - \bar{\alpha}_k} \epsilon,
    \end{equation*}
    and the diffusion model predicts \(\hat{\epsilon} = \epsilon_\theta(x_k, k, x_{hist})\), conditioned on \(x_{hist}\). We model the conditional likelihood using the EBM in Equation~\eqref{eq:ebm_dist} which is equivalent to:
    
    \begin{equation*}
        p_{\theta, \phi}(x | \epsilon, x_{hist}) = \frac{1}{Z_{\theta, \phi}(\epsilon, x_{hist})} \exp\left( - \tilde{E}_{\theta, \phi}(x, \epsilon, x_{hist}) \right),
    \end{equation*}
    with the energy function:
    
    \begin{equation}
        \tilde{E}_{\theta, \phi}(x, \epsilon, x_{hist}) = E_\phi(\hat{\epsilon}) + \frac{1}{2\sigma^2} \|\hat{\epsilon} - \epsilon\|_2^2,
        \label{eq:energy_function}
    \end{equation}
    where \(\hat{\epsilon} = \epsilon_\theta(x_k, k, x_{hist})\), \(\sigma^2\) is a variance hyperparameter, and \(Z_{\theta, \phi}(\epsilon, x_{hist}) = \int \exp\left( - \tilde{E}_{\theta, \phi}(x', \epsilon, x_{hist}) \right) dx'\) is the normalization constant. The term \(\frac{1}{2\sigma^2} \|\hat{\epsilon} - \epsilon\|_2^2\) ensures \(\hat{\epsilon} \approx \epsilon\) for normal data, while \(E_\phi(\hat{\epsilon})\) penalizes deviations from normality given \(x_{hist}\).
    
    Substituting into the marginal likelihood:
    
    \begin{align}
        p_{\theta, \phi}(x | x_{hist}) &= \int \mathcal{N}(\epsilon | 0, I) \cdot \frac{1}{Z_{\theta, \phi}(\epsilon, x_{hist})} \notag \\
        &\quad \cdot \exp\left( - E_\phi(\hat{\epsilon}) - \frac{1}{2\sigma^2} \|\hat{\epsilon} - \epsilon\|_2^2 \right) d\epsilon.
        \label{eq:subst_marginal_likelihood}
    \end{align}
    
    % This integral is intractable due to the non-linear \(\hat{\epsilon} = \epsilon_\theta(x_k, k, x_{hist})\), \(E_\phi(\hat{\epsilon})\), and the \(\epsilon\)- and \(x_{hist}\)-dependent \(Z_{\theta, \phi}(\epsilon, x_{hist})\). 
	The integral is intractable because the partition function \(Z_{\theta,\phi}(\epsilon,x_{\text{hist}})\) depends on \(\epsilon\) and \(x_{\text{hist}}\), and this dependence interacts with the nonlinear mappings \(\hat{\epsilon}=\epsilon_\theta(x_k,k,x_{\text{hist}})\) and \(E_\phi(\hat{\epsilon})\).
	% , preventing a closed form.
	% The integral is intractable because $Z_{\theta, \phi}(\epsilon, x_{\text{hist}})$, which is dependent on $\epsilon$ and $x_{\text{hist}}$, complicates the non-linear terms $\hat{\epsilon} = \epsilon_\theta(x_k, k, x_{\text{hist}})$ and $E_\phi(\hat{\epsilon})$.
	We approximate it using a single Monte Carlo sample \(\epsilon \sim \mathcal{N}(0, I)\):
    
    \begin{align}
        \log p_{\theta, \phi}(x | x_{hist}) &\approx \log \mathcal{N}(\epsilon | 0, I) - E_\phi(\hat{\epsilon}) \notag \\
        &\quad - \frac{1}{2\sigma^2} \|\hat{\epsilon} - \epsilon\|_2^2 - \log Z_{\theta, \phi}(\epsilon, x_{hist}).
        \label{eq:approx_log_likelihood}
    \end{align}
    
    Since \(\log \mathcal{N}(\epsilon | 0, I) = -\frac{d}{2} \log (2\pi) - \frac{1}{2} \|\epsilon\|_2^2\), we have:
    
    \begin{align}
        \log p_{\theta, \phi}(x | x_{hist}) &\approx -\frac{d}{2} \log (2\pi) - \frac{1}{2} \|\epsilon\|_2^2 - E_\phi(\hat{\epsilon}) \notag \\
        &\quad - \frac{1}{2\sigma^2} \|\hat{\epsilon} - \epsilon\|_2^2 - \log Z_{\theta, \phi}(\epsilon, x_{hist}).
        \label{eq:expanded_log_likelihood}
    \end{align}
    
    For normal data conditioned on \(x_{hist}\), \(\hat{\epsilon} \approx \epsilon\) (from diffusion model training), so \(\|\hat{\epsilon} - \epsilon\|_2^2 \approx 0\), and \(E_\phi(\hat{\epsilon})\) is small due to EBM training. 
	% For anomalies, \(\hat{\epsilon}\) deviates, increasing the volatility of both  terms. 
	For anomalies, $\hat{\epsilon}$ deviates, increasing both the squared L2 norm $\|\hat{\epsilon} - \epsilon\|_2^2$ and the energy score $E_\phi(\hat{\epsilon})$.
	Since \(Z_{\theta, \phi}(\epsilon, x_{hist})\) is difficult to compute and relatively constant across samples, we focus on the dominant term for anomaly scoring:
    \begin{equation}
        \log p_{\theta, \phi}(x | x_{hist}) \approx - E_\phi(\hat{\epsilon}) + \text{const},
        \label{eq:simplified_log_likelihood}
    \end{equation}
    yielding the anomaly score:
    \begin{equation}
        s(x) = E_\phi(\hat{\epsilon}), \quad \hat{\epsilon} = \epsilon_\theta(x_k, k, x_{hist}).
        \label{eq:anomaly_score}
    \end{equation}
    To ensure \(E_\phi(\hat{\epsilon})\) distinguishes normal from anomalous data, the EBM is trained with the contrastive loss:
    \begin{align}
        \mathcal{L}_{\text{EBM}} = \mathbb{E}_{x \sim p_{\text{data}}(\cdot | x_{hist})} \left[ E_\phi(\hat{\epsilon}^+) \right] - \mathbb{E}_{\hat{\epsilon}^- \sim p_{\text{neg}}} \left[ E_\phi(\hat{\epsilon}^-) \right],
        \label{eq:ebm_loss}
    \end{align}
    where \(\hat{\epsilon}^+\) comes from normal data conditioned on \(x_{hist}\), and \(\hat{\epsilon}^-\) from anomalous synthetic samples via the forward diffusion. 
	\textcolor{black}{This validates Proposition~\ref{prop:ebm_score} by showing that \(E_\phi(\hat{\epsilon})\) is a proxy for \(-\log p_{\theta,\phi}(x \mid x_{\text{hist}})\) and the high scores indicate anomalies.}
	% This validates Proposition~\ref{prop:ebm_score}, as \(E_\phi(\hat{\epsilon})\) serves as a proxy for \(-\log p_{\theta, \phi}(x | x_{hist})\), with high scores indicating anomalies.
    
    \section{Justification for One-Step Diffusion and Noise Prediction in Anomaly Detection}
    \label{app:justification_single_step}

    This appendix justifies the use of diffusing the original data for one \textcolor{black}{time step} and predicting the noise in our diffusion-based anomaly detection framework, conditioned on historical data \(x_{hist}\). We combine empirical evidence from prior work on diffusion models with a mathematical derivation to demonstrate its advantages over direct noise prediction from the original data, enhancing our \textcolor{black}{justifications}.
    
    The previous studies on diffusion models~\citep{ho2020} suggest that predicting the noise added during the diffusion process outperforms other parameterizations, such as predicting the mean of the reverse process or the original data. This approach aligns with generative modeling principles and improves the quality of generated samples. 
	For anomaly detection, accurately modeling the normal conditional \(p(x \mid x_{hist})\) is also key to detecting deviations. The empirical advantage therefore supports predicting noise rather than alternative targets.
	% For anomaly detection, where accurately modeling the conditional distribution of normal data given historical context \(x_{hist}\) is critical to identifying deviations, this empirical advantage supports our choice to predict noise rather than alternative targets.
    
    Considering a test sample \(x\), we diffuse \(x\) for one time step (\(k=1\)) using the forward process in Equation~\eqref{eq:forward_diffusion} as follows:
    \begin{equation*}
        x_1 = \sqrt{\bar{\alpha}_1} x + \sqrt{1 - \bar{\alpha}_1} \epsilon, \quad \epsilon \sim \mathcal{N}(0, I),
    \end{equation*}
    where \(\bar{\alpha}_1 = 1 - \beta_1\), and \(\beta_1\) is a small noise variance. The diffusion model, parameterized by \(\epsilon_\theta\), predicts the noise \(\hat{\epsilon} = \epsilon_\theta(x_1, 1, x_{hist})\), conditioned on \(x_{hist}\), and is trained on normal data \(x_0 \sim p_{\text{data}}(\cdot | x_{hist})\) to minimize Equation~\eqref{eq:diffusion_loss} as follows:
    \begin{equation*}
        \mathcal{L}_{\text{DM}} = \mathbb{E}_{x_0 \sim p_{\text{data}}(\cdot | x_{hist}), \epsilon \sim \mathcal{N}(0, I), k} \left[ \|\epsilon - \epsilon_\theta(x_k, k, x_{hist})\|_2^2 \right].
    \end{equation*}
    Our goal is to detect anomalies by analyzing \(\hat{\epsilon}\), leveraging its connection to the conditional data distribution.
    
    % limitation of direct prediction
    First, consider predicting noise directly from the original data without diffusion (\(k=0\)). Here, \(x_0 = x\), and \(\bar{\alpha}_0 = 1\), so no noise is added. The predicted noise becomes:
    \begin{align}
        \epsilon_\theta(x_0, 0, x_{hist}) &\approx -\sqrt{1 - \bar{\alpha}_0} \nabla_{x_0} \log p_0(x_0 | x_{hist}) \notag \\
        &= 0 \cdot \nabla_{x_0} \log p_{\text{data}}(x | x_{hist}) = 0,
        \label{eq:direct_prediction}
    \end{align}
    as derived in Proposition~\ref{prop:noise_score}. This result is trivial and independent of \(x\), offering no discriminative power for anomaly detection. Thus, direct prediction fails to provide meaningful information about the data’s conformity to \(p_{\text{data}}(x | x_{hist})\).
    
    Now, diffusing for one step (\(k=1\)) introduces a small perturbation. For normal data \(x \sim p_{\text{data}}(\cdot | x_{hist})\), the model is trained such that \(\epsilon_\theta(x_1, 1, x_{hist}) \approx \epsilon\), where \(\epsilon \sim \mathcal{N}(0, I)\). Define the predicted noise distribution for normal data:
    \begin{equation}
        p_{\hat{\epsilon}}^{\text{normal}}(\hat{\epsilon}) = \mathbb{E}_{x \sim p_{\text{data}}(\cdot | x_{hist}), \epsilon \sim \mathcal{N}(0, I)} \left[ \delta(\hat{\epsilon} - \epsilon_\theta(x_1, 1, x_{hist})) \right].
        \label{eq:noise_distribution}
    \end{equation}
    Since \(\epsilon_\theta\) minimizes the mean squared error to \(\epsilon\), \(p_{\hat{\epsilon}}^{\text{normal}}(\hat{\epsilon}) \approx \mathcal{N}(0, I)\). For anomalous data \(x \not\sim p_{\text{data}}(\cdot | x_{hist})\), \(x_1\) lies outside the manifold learned during training, and \(\epsilon_\theta(x_1, 1, x_{hist})\) deviates from \(\mathcal{N}(0, I)\), reflecting a mismatch with the expected noise pattern given \(x_{hist}\).
    
    % Quantifying the deviation
    To formalize this, consider the expected norm of the predicted noise. For normal data:
    \begin{equation}
        \mathbb{E}_{x \sim p_{\text{data}}(\cdot | x_{hist})} \left[ \|\epsilon_\theta(x_1, 1, x_{hist})\|_2^2 \right] \approx \mathbb{E}_{\epsilon \sim \mathcal{N}(0, I)} \left[ \|\epsilon\|_2^2 \right] = d,
        \label{eq:expected_norm_normal}
    \end{equation}
    where \(d\) is the data dimension. For anomalous data, let \(x \sim p_{\text{anom}}\). Since \(\epsilon_\theta\) is not trained on \(p_{\text{anom}}\), \(\hat{\epsilon}\) may exhibit larger variance or bias. Suppose \(\epsilon_\theta(x_1, 1, x_{hist}) \sim \mathcal{N}(\mu_{\text{anom}}, \Sigma_{\text{anom}})\), where \(\mu_{\text{anom}} \neq 0\) or \(\Sigma_{\text{anom}} \neq I\). Then:
    \begin{equation}
        \mathbb{E}_{x \sim p_{\text{anom}}} \left[ \|\epsilon_\theta(x_1, 1, x_{hist})\|_2^2 \right] = \text{tr}(\Sigma_{\text{anom}}) + \|\mu_{\text{anom}}\|_2^2,
        \label{eq:expected_norm_anom}
    \end{equation}
    which typically exceeds \(d\) due to distributional mismatch. This deviation enables anomaly detection by thresholding \(\|\hat{\epsilon}\|_2^2\).
    
    From Proposition~\ref{prop:noise_score}, \(\epsilon_\theta(x_1, 1, x_{hist}) \approx -\sqrt{1 - \bar{\alpha}_1} \nabla_{x_1} \log p_1(x_1 | x_{hist})\). For small \(k\), \(p_1(x_1 | x_{hist}) \approx p_{\text{data}}(x | x_{hist})\), so:
    \begin{equation}
        \hat{\epsilon} \approx -\sqrt{1 - \bar{\alpha}_1} \nabla_x \log p_{\text{data}}(x | x_{hist}).
        \label{eq:score_connection}
    \end{equation}
    The score function’s magnitude is larger in low-density regions, \textcolor{black}{which are typical {for} anomalies}. Thus, \(\|\hat{\epsilon}\|_2\) is larger for anomalous \(x\), providing a theoretical basis for detection.  
	\textcolor{black}{In practice, we use the full distribution of \(\hat{\epsilon}\) for robustness.}
	% , though we use the full distribution of \(\hat{\epsilon}\) for robustness.
    Diffusing for one step and predicting noise conditioned on \(x_{hist}\) offers:
    \begin{enumerate}
        \item {Non-trivial signal}: Unlike \(k=0\), where \(\hat{\epsilon} = 0\), \(k=1\) yields a meaningful \(\hat{\epsilon}\) reflecting the conditional data distribution.
        \item {Distributional sensitivity}: \(\hat{\epsilon}\) approximates \(\mathcal{N}(0, I)\) for normal data but deviates for anomalies, enabling detection given \(x_{hist}\).
        \item {Score function insight}: \(\hat{\epsilon}\) encodes the conditional score, linking to density-based anomaly principles.
        \item {Efficiency}: A single diffusion step and prediction suffice, ideal for real-time applications.
    \end{enumerate}
    % This approach leverages diffusion models’ strengths, providing a solid foundation for anomaly detection in our framework, enhanced by historical context.
	\section{Evaluation Metrics Details}
	\label{app:evaluation_metrics}

	\begin{itemize}
		\item {Precision}: Proportion of true positives (TP) among all positive predictions:
			  \begin{equation*}
				  \text{Precision} = \frac{\text{TP}}{\text{TP} + \text{FP}}.
			  \end{equation*}
		\item {Recall}: Proportion of TP among all actual positives:
			  \begin{equation*}
				  \text{Recall} = \frac{\text{TP}}{\text{TP} + \text{FN}}.
			  \end{equation*}
		\item {F1-score}: Harmonic mean of precision and recall:
			  \begin{equation*}
				  F1 = 2 \cdot \frac{\text{Precision} \cdot \text{Recall}}{\text{Precision} + \text{Recall}}.
			  \end{equation*}
		\item {Accuracy}: Proportion of correctly classified samples:
			  \begin{align*}
				  \text{Accuracy} = \frac{\text{TP} + \text{TN}}{\text{TP} + \text{TN} + \text{FP} + \text{FN}}.
			  \end{align*}
	\end{itemize}

	To account for the unique nature of UAV faults, such as engine failures that occur once and persist until the flight ends, we adjust these metrics using the affiliation approach~\citep{huet2022}. Classical metrics punish predictions slightly off the fault’s start (e.g., a second late) and overvalue long faults, skewing scores. Affiliation metrics measure the time gap between predictions and the fault’s start, comparing to a random guess to compute precision and recall. 
	\textcolor{black}{It} ensures: (1) predictions close to the fault’s onset (e.g., detecting engine failure within seconds) are rewarded; (2) the fault is scored as one event, avoiding bias from its length; and (3) only predictions near the actual fault improve the score, preventing unrelated guesses (e.g., random alerts far from the fault) from falsely boosting results. For image data (CIFAR-10 dataset), to align with baselines like NeuTraL-F and PANDA (Table V), we use accuracy, ensuring fair comparison with established image anomaly detection methods.

    \section{Datasets Statistics}\label{app:dataset_stats}
	\begin{table*}[t]
		\centering
		\footnotesize
		\caption{Fault types in the processed ALFA dataset, detailing test cases and flight time.}
		\label{tab:fault_types}
		% \resizebox{\textwidth}{!}{
		\begin{tabular}{@{}lrrr@{}}
		\toprule
		\textbf{Fault Category} & \textbf{Test Flights} & \textbf{Before Fault (s)} & \textbf{With Fault (s)} \\
		\midrule
		Engine full power loss & 23 & 2282 & 362 \\
		Rudder stuck to left & 1 & 60 & 9 \\
		Rudder stuck to right & 2 & 107 & 32 \\
		Elevator stuck at zero & 2 & 181 & 23 \\
		Left aileron stuck at zero & 3 & 228 & 183 \\
		Right aileron stuck at zero & 4 & 442 & 231 \\
		Both ailerons stuck at zero & 1 & 66 & 36 \\
		Rudder and aileron at zero \& Aileron Zero & 1 & 116 & 27 \\
		Total & 47 & 3935 & 777 \\
		\bottomrule
		\end{tabular}
		% }
		\end{table*}

		\begin{table*}[h]
			\centering
			\footnotesize
			\caption{SMD and CIFAR-10 Datasets.}
			\label{tab:raw_data_stats}
			% \resizebox{\textwidth}{!}{
			\begin{tabular}{@{}lcccc@{}}
			\toprule
			\textbf{Dataset} & \textbf{Train} & \textbf{Test} & \textbf{Dimensions} & \textbf{Anomalies (\%)} \\
			\midrule
			SMD              & 708,405        & 708,420       & 38 (4 traces)                 & 4.16                    \\
			CIFAR-10         & 50,000         & 10,000        & 3,072 (32$\times$32$\times$3) & 90.00 (test)            \\
			\bottomrule
			\end{tabular}
			% }
			\end{table*}

			\begin{table*}[t]
				\centering
				\footnotesize
				\caption{Sensor Descriptions for ALFA Dataset.}
				\label{tab:sensors}
				% \resizebox{\textwidth}{!}{
				\begin{tabular}{@{}clp{6cm}@{}}
				\toprule
				\textbf{Index} & \textbf{Sensor Name} & \textbf{Description} \\
				\midrule
				0              & mavros\_nav\_info\_errors               & Tracking, airspeed, and altitude errors                              \\
				1              & mavros\_vfr\_hud                        & Data for HUD (climb, altitude, groundspeed, heading, throttle)       \\
				2              & mavros\_local\_position\_odom           & Local position and odometry (angular/linear velocities, orientation) \\
				3              & mavros\_imu\_data                       & IMU state (angular velocity, linear acceleration, orientation)       \\
				4              & mavros\_wind\_estimation                & Wind estimation by FCU (wind speed components)                       \\
				5              & mavros\_global\_position\_raw\_gps\_vel & GPS-based velocity (linear velocities)                               \\
				6              & mavros\_nav\_info\_velocity             & Commanded and measured velocity                                      \\
				7              & mavros\_rc\_out                         & Remote control outputs (throttle, aileron channels)                  \\
				8              & mavros\_global\_position\_global        & Global position info (altitude, longitude, latitude)                 \\
				9              & mavros\_imu\_mag                        & Magnetic field components                                            \\
				10             & mavros\_setpoint\_raw\_target\_global   & Setpoint messages (acceleration or force setpoints)                  \\
				11             & mavctrl\_path\_dev                      & Path deviation                               \\
				12             & mavros\_nav\_info\_yaw                  & Commanded and measured yaw                                           \\
				13             & mavros\_nav\_info\_pitch                & Commanded and measured pitch                                         \\
				14             & mavros\_global\_position\_rel\_alt      & Relative altitude                                                    \\
				15             & mavctrl\_rpy                            & Measured roll, pitch, and yaw                                        \\
				16             & mavros\_global\_position\_compass\_hdg  & Compass heading                                                      \\
				17             & mavros\_nav\_info\_roll                 & Commanded and measured roll                                          \\
				18             & mavros\_imu\_atm\_pressure              & Atmospheric pressure                                                 \\
				\bottomrule
				\end{tabular}
				% }
				\end{table*}
    
    Table~\ref{tab:fault_types} provides details on the fault types in the processed ALFA dataset, including the number of test cases and flight times before and after faults. Table~\ref{tab:raw_data_stats} summarizes the SMD and CIFAR-10 datasets used in our experiments, including the number of training and test samples, dimensions, and anomaly percentages.

    \section{Sensor Descriptions fo ALFA Dataset}
    \label{app:sensor_descriptions}
    Table~\ref{tab:sensors} lists the sensors used in the ALFA dataset after we processed the raw data and converted it into a graph strucutre for our experiments. Each sensor is indexed and described, providing insights into the UAV's operational parameters and environmental conditions during flights.

	\section{Additional Visualizations of Experiments}
	\label{app:additional_visualizations}

\begin{figure*}[t]
	\centering
	\begin{minipage}[t]{0.48\textwidth}
	  \centering
	  \captionsetup[subfloat]{font=tiny,labelfont=tiny}
	  \subfloat[DM-NP (KDE) Scores (BASiC)]{
		\includegraphics[width=0.99\linewidth]{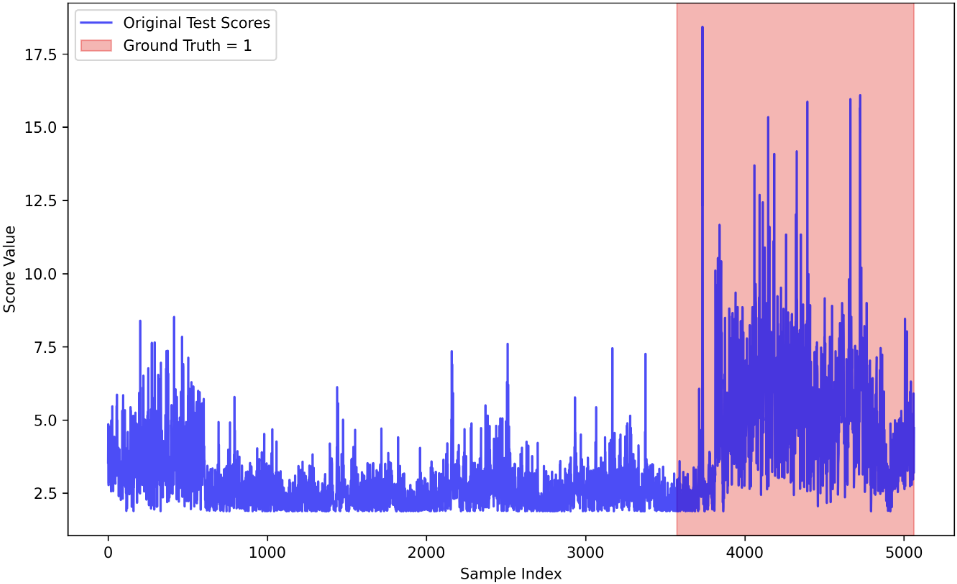}
		\label{fig:dmkde_scores_basic}
	  }
	  \hfill
	  \subfloat[DM-P Scores (BASiC)]{
		\includegraphics[width=0.99\linewidth]{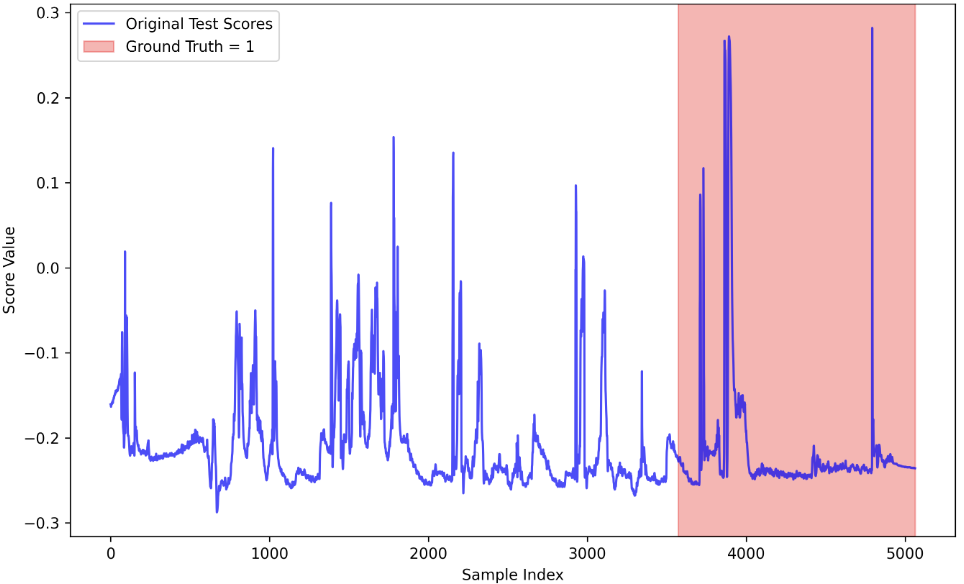}
		\label{fig:dmebm_scores_basic}
	  }
	\end{minipage}
	\hfill
	\begin{minipage}[t]{0.48\textwidth}
	  \centering
	  \captionsetup[subfloat]{font=tiny,labelfont=tiny}
	  \subfloat[DM-NP (KDE) Predictions (BASiC)]{
		\includegraphics[width=0.99\linewidth]{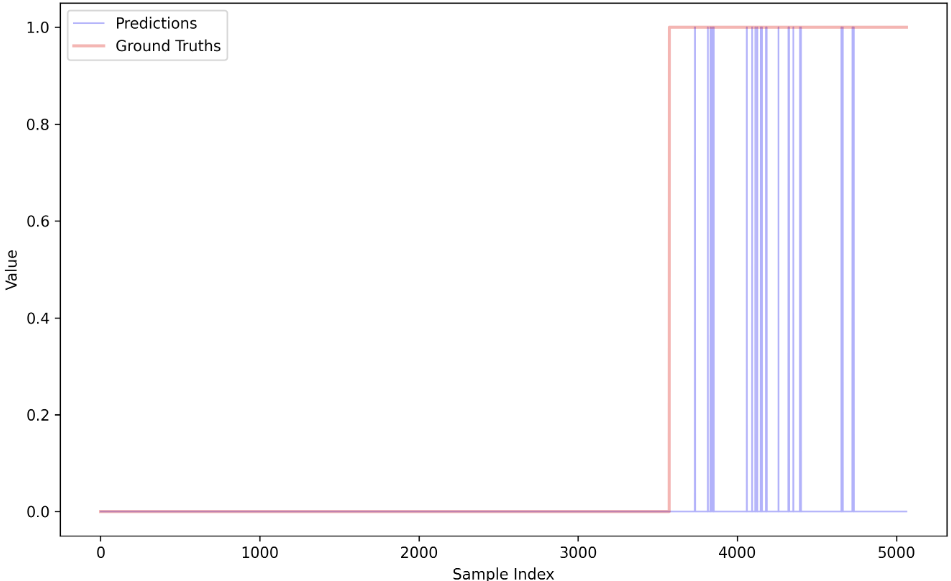}
		\label{fig:dmkde_predictions_basic}
	  }
	  \hfill
	  \subfloat[DM-P Predictions (BASiC)]{
		\includegraphics[width=0.99\linewidth]{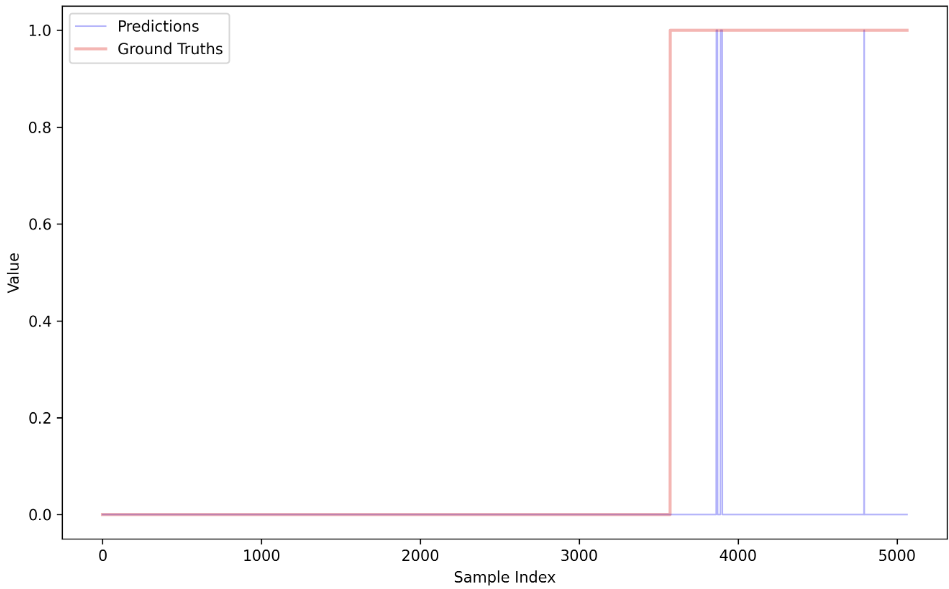}
		\label{fig:dmebm_predictions_basic}
	  }
	\end{minipage}
	\caption{BASiC: anomaly scores (left) and predictions (right). Ground-truth fault regions are shaded in red.}
	\label{fig:combined_results_basic_only}
  \end{figure*}

  \begin{figure*}[t]
	\centering
	\begin{minipage}[t]{0.48\textwidth}
	  \centering
	  \captionsetup[subfloat]{font=tiny,labelfont=tiny}
	  % SMD 2-1 scores
	  \subfloat[DM-NP (KNN) Scores (SMD 2-1)]{
		\includegraphics[width=0.99\linewidth]{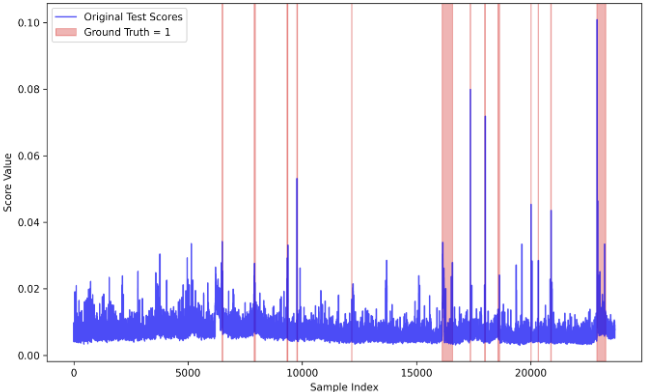}
		\label{fig:dmknn_scores_smd21}
	  }
	  \hfill
	  \subfloat[DM-P Scores (SMD 2-1)]{
		\includegraphics[width=0.99\linewidth]{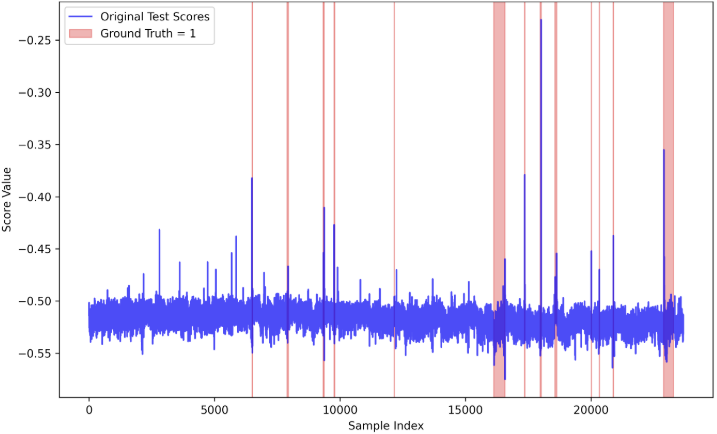}
		\label{fig:dmebm_scores_smd21}
	  }
	  \hfill
	  \subfloat[DM-NP (KDE) Scores (SMD 1-1)]{
		\includegraphics[width=0.99\linewidth]{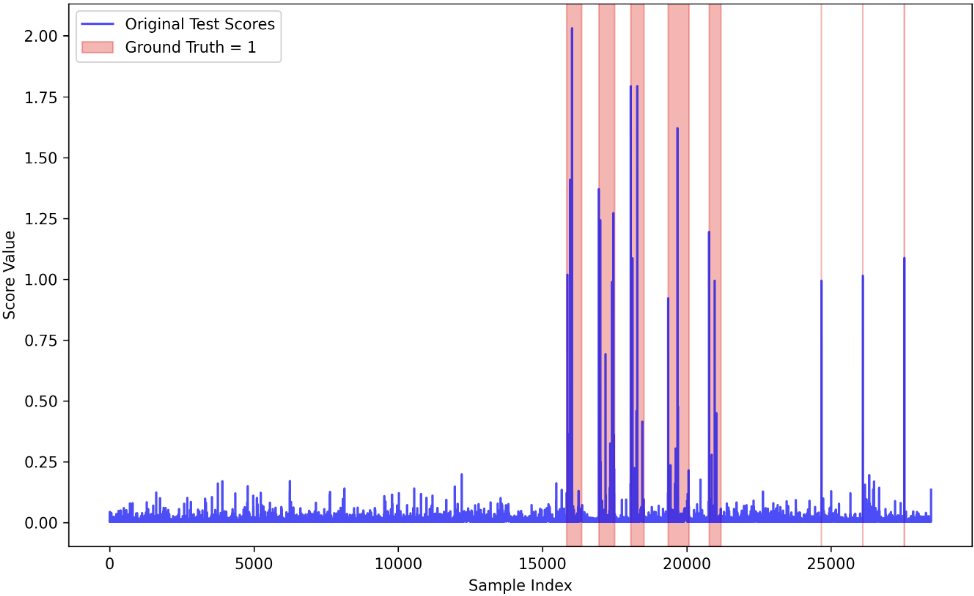}
		\label{fig:dmkde_scores_smd11}
	  }
	  \hfill
	  \subfloat[DM-P Scores (SMD 1-1)]{
		\includegraphics[width=0.99\linewidth]{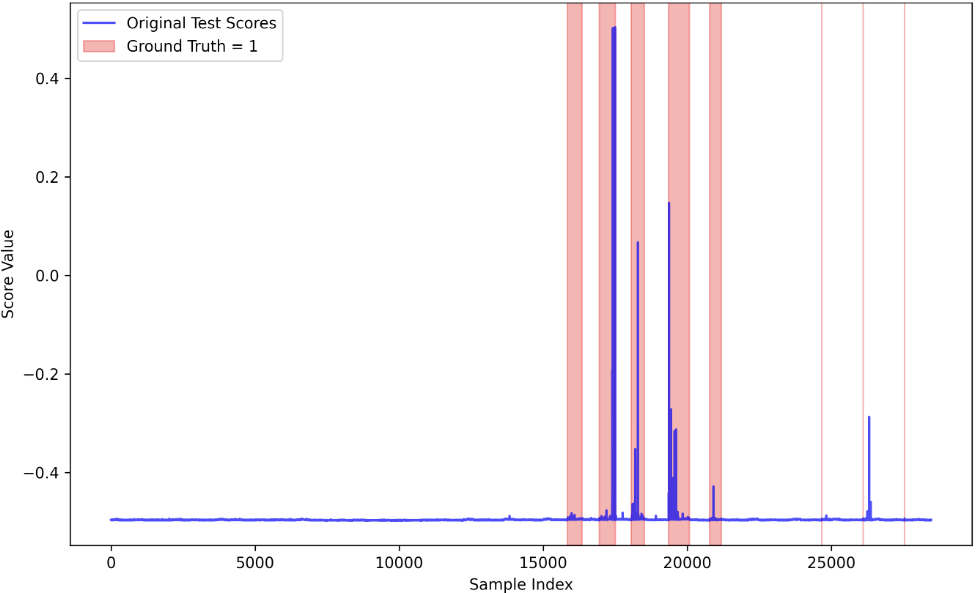}
		\label{fig:dmebm_scores_smd11}
	  }
	\end{minipage}
	\hfill
	\begin{minipage}[t]{0.48\textwidth}
	  \centering
	  \captionsetup[subfloat]{font=tiny,labelfont=tiny}
	  \subfloat[DM-NP (KNN) Predictions (SMD 2-1)]{
		\includegraphics[width=0.98\linewidth]{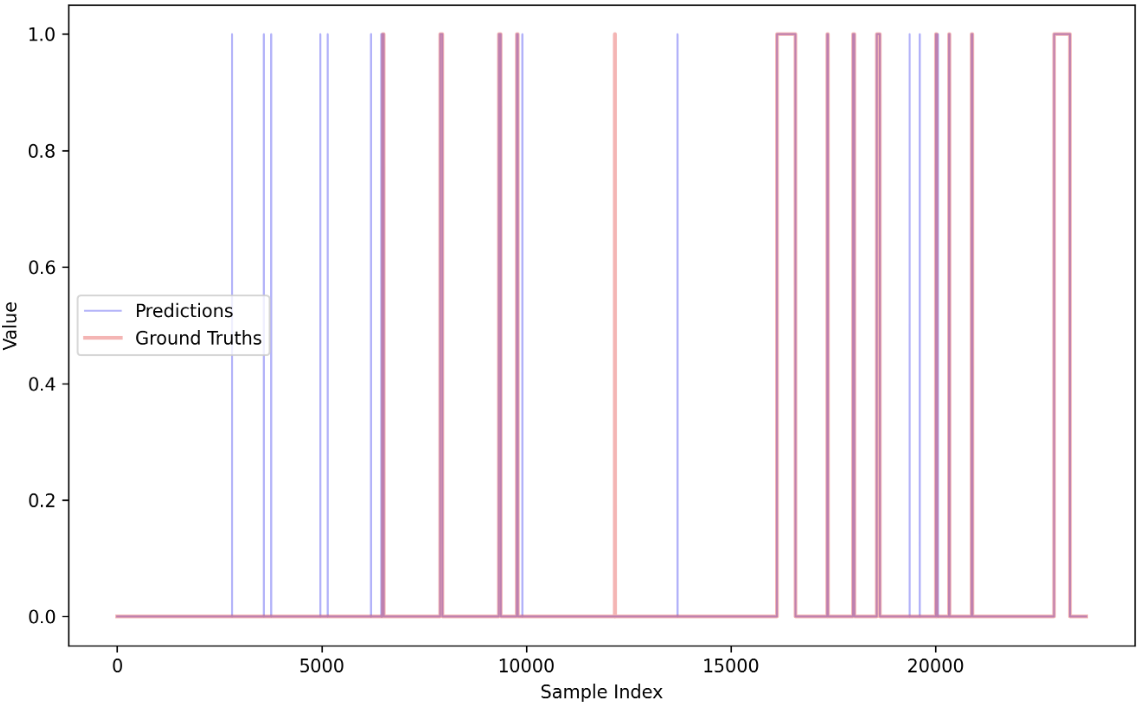}
		\label{fig:dmknn_predictions_smd21}
	  }
	  \hfill
	  \subfloat[DM-P Predictions (SMD 2-1)]{
		\includegraphics[width=0.97\linewidth]{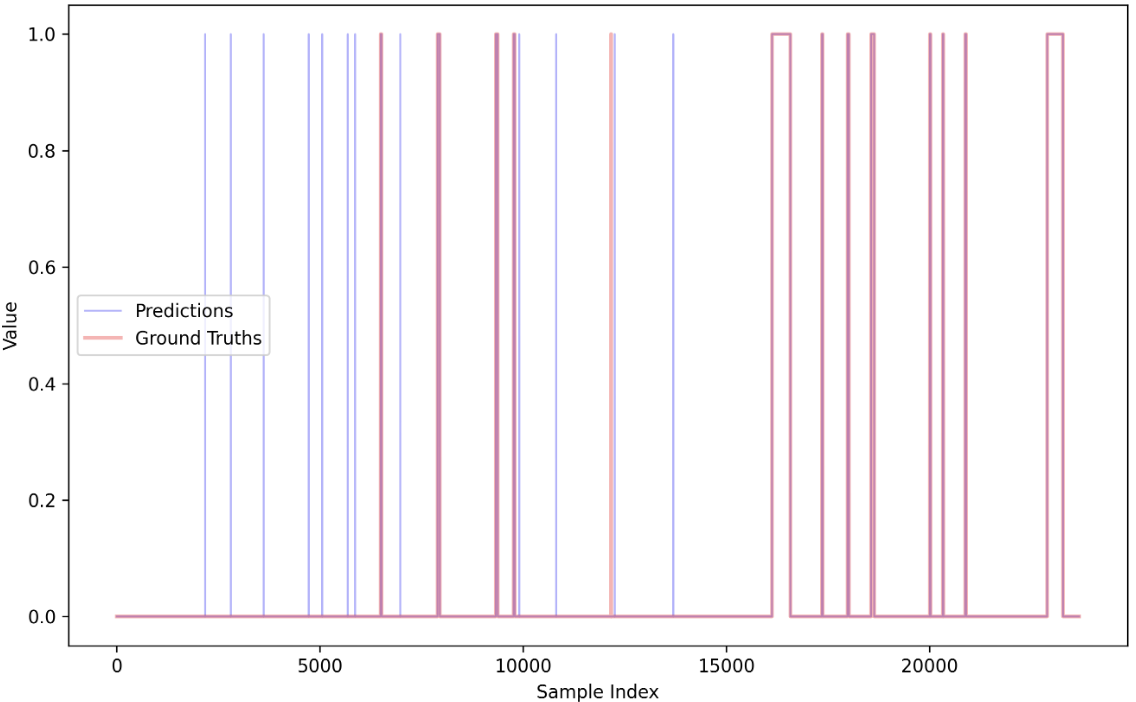}
		\label{fig:dmebm_predictions_smd21}
	  }
	  \hfill
	  \subfloat[DM-NP (KDE) Predictions (SMD 1-1)]{
		\includegraphics[width=0.97\linewidth]{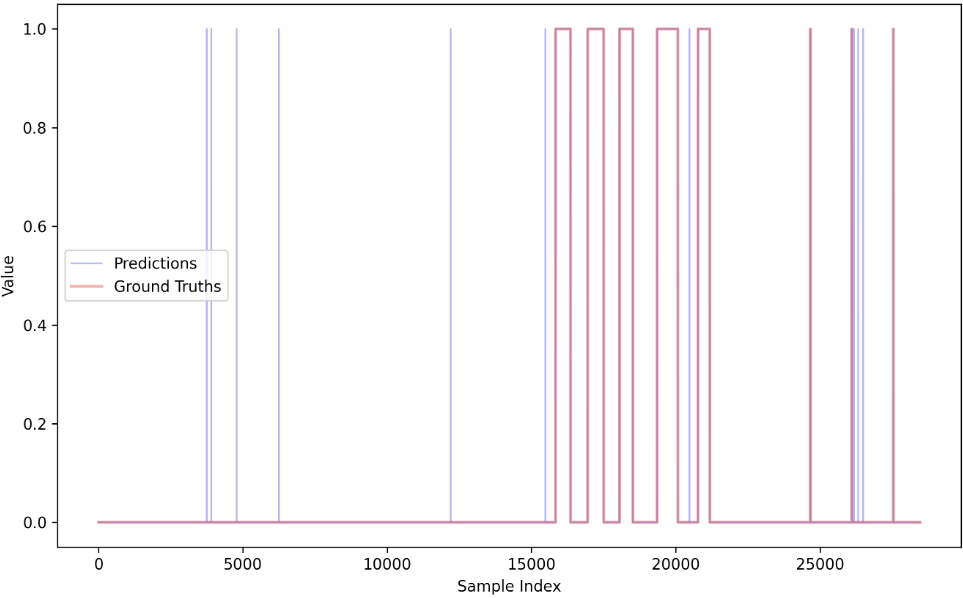}
		\label{fig:dmkde_predictions_smd11}
	  }
	  \hfill
	  \subfloat[DM-P Predictions (SMD 1-1)]{
		\includegraphics[width=0.97\linewidth]{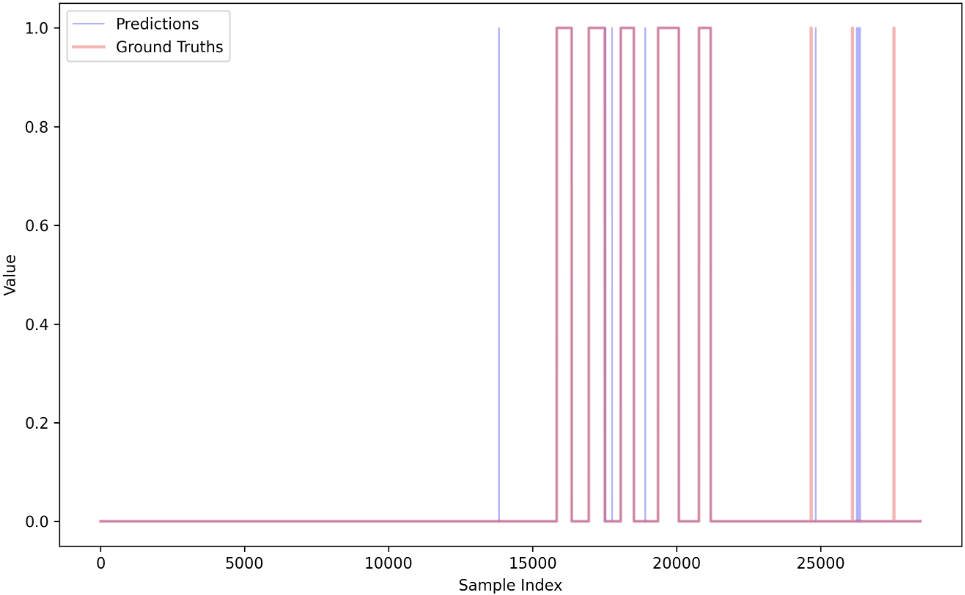}
		\label{fig:dmebm_predictions_smd11}
	  }
	\end{minipage}
	\caption{SMD: anomaly scores (left) and predictions (right). Top row: Machine~2-1. Bottom row: Machine~1-1. Ground-truth fault regions are shaded in red.}
	\label{fig:combined_results_SMD_both}
  \end{figure*}

	\textcolor{black}{Figures~\ref{fig:combined_results_basic_only} and \ref{fig:combined_results_SMD_both} provide additional visualizations of anomaly scores and predictions on the BASIC and SMD datasets from both branches (DM-NP and DM-P). The left column shows scores and the right column shows the corresponding predictions. On BASIC (graph-structured UAV data), we observe zero false positives. On SMD datasets (pure time series), we almost miss no true positives. These results exceed recent baselines reported in the main paper.}

	\section{Gradient Dynamics and Trajectory Analysis of DM-P} \label{app:gradient_dynamics}

\begin{figure*}[h]
	\centering
	% Single column: SMD dataset visualizations (2x2 grid)
	\begin{minipage}[t]{\textwidth}
		\centering
		\captionsetup[subfloat]{font=tiny,labelfont=tiny} 
		\subfloat[Original Data Distribution after PCA (SMD)]{
			\includegraphics[width=0.45\linewidth]{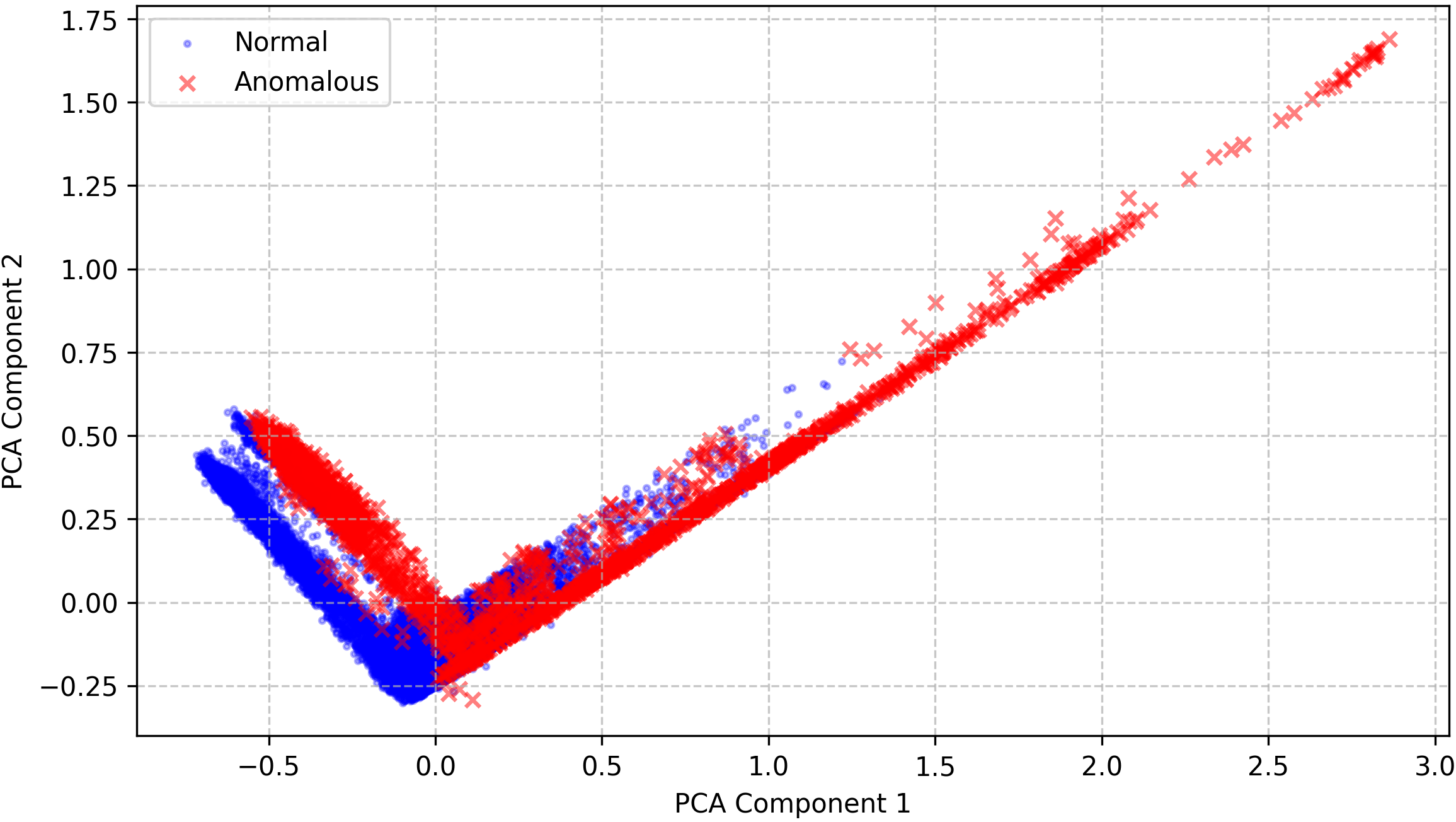}
			\label{fig:dmebm_vis_a_smd}
		}
		\hfill
		\subfloat[Diffused Data at \( t = 1 \) with Added Noise (SMD)]{
			\includegraphics[width=0.45\linewidth]{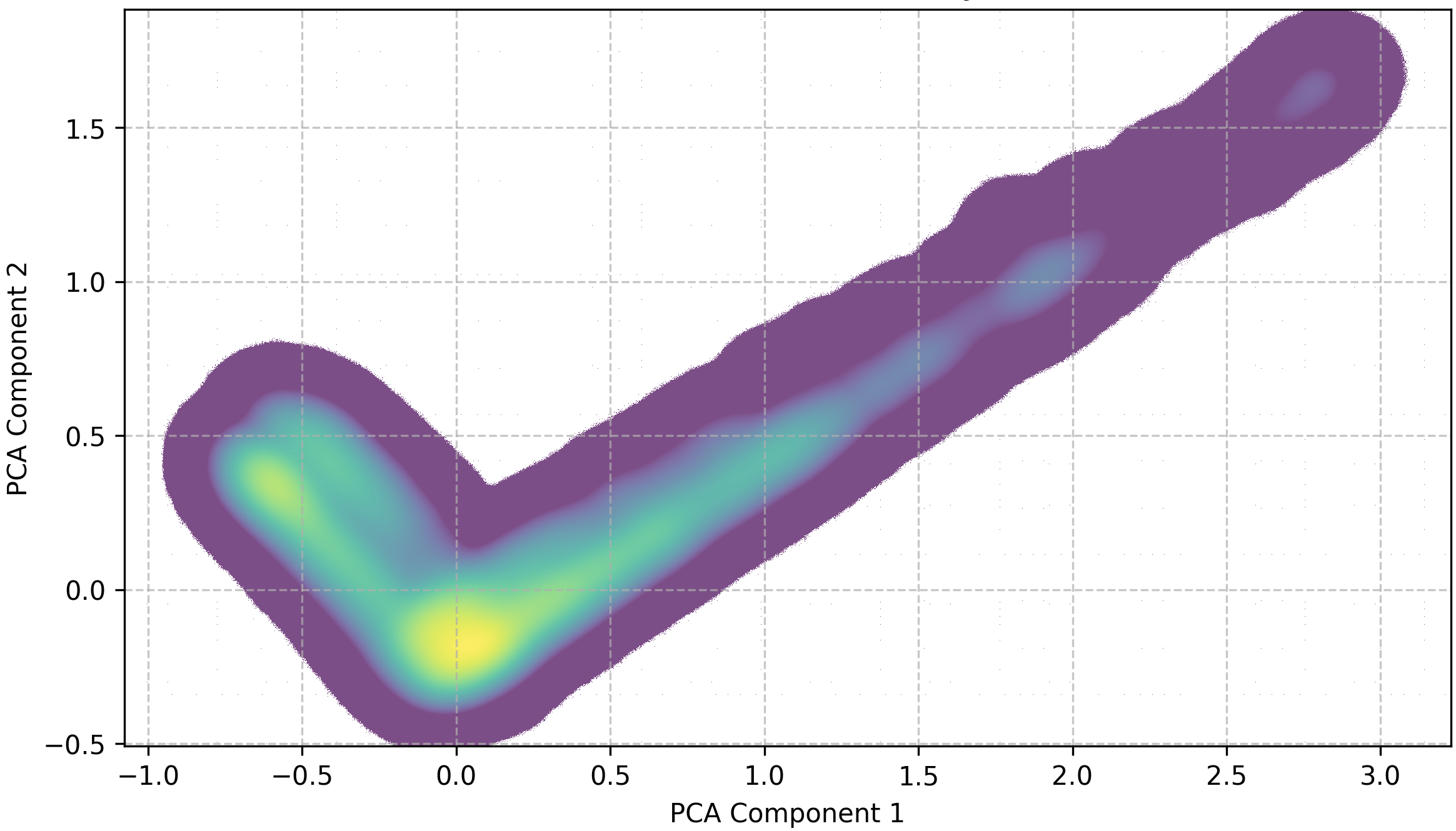}
			\label{fig:dmebm_vis_b_smd}
		}

		\subfloat[DM-P (EBM) Energy Gradient Field at \( t = 1 \) (SMD)]{
			\includegraphics[width=0.45\linewidth]{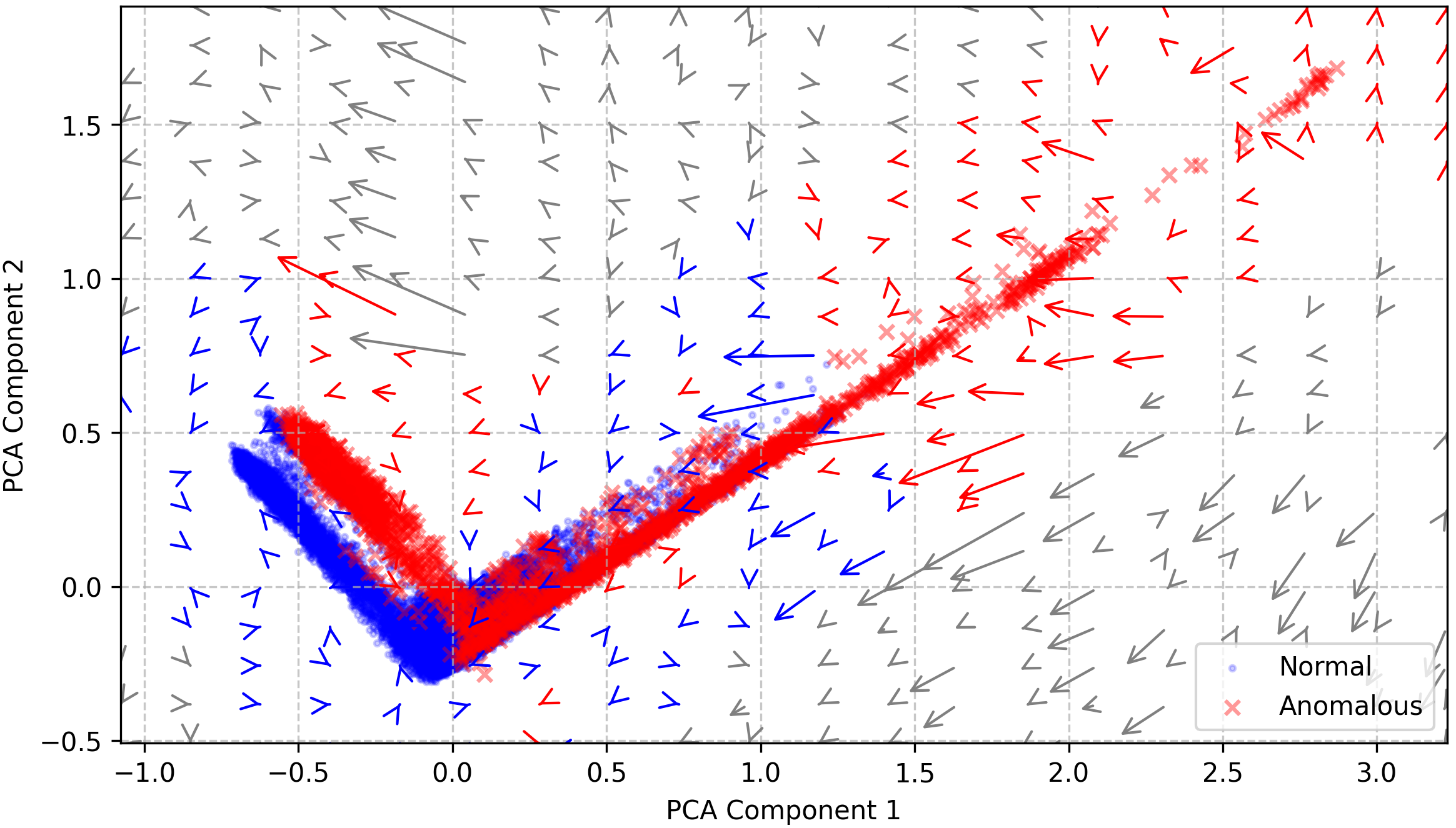}
			\label{fig:dmebm_vis_c_smd}
		}
		\hfill
		\subfloat[Data Evolution Trajectories under DM-P (SMD)]{
			\includegraphics[width=0.45\linewidth]{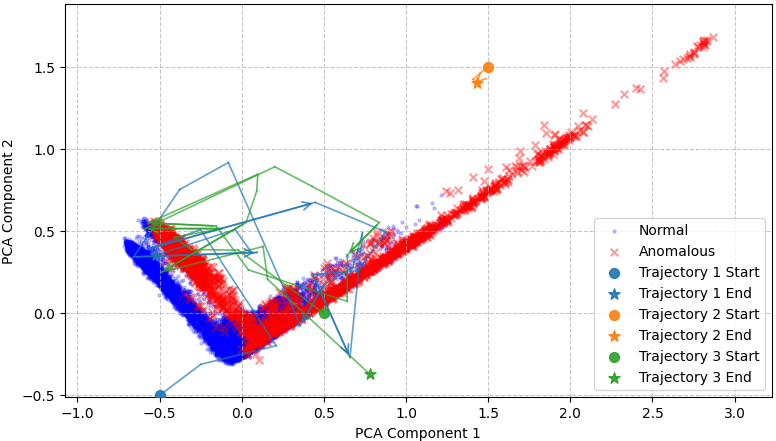}
			\label{fig:dmebm_vis_f_smd}
		}
	\end{minipage}
	\caption{DM-P Framework Analysis on SMD Dataset (Machine 1-1). Subfigures (a--d) depict: (a) original data distribution after PCA, (b) diffused data at time step \( k = 1 \), (c) energy gradient field at \( t = 1 \), and (d) data evolution trajectories.}
	\label{fig:vis_dtdp}
\end{figure*}

To further evaluate the effectiveness of the DM-P framework, parameterized by an Energy-Based Model (EBM), we present a visualization analysis using the SMD dataset in Figure~\ref{fig:vis_dtdp}. Figure~\ref{fig:vis_dtdp} (a) depicts the original data distribution after PCA, with normal data (blue dots) clustering densely around 0.0 on the \emph{x}-axis and anomalous data (red `x' markers) scattered along a diagonal from 0.0 to 3.0, with notable overlap in the lower left quadrant that complicates the anomaly detection task. Figure~\ref{fig:vis_dtdp} (b) shows the data after one diffusion time step ($ k = 1 $), where a density plot (purple to yellow gradient) highlights the densest region around 0.0, preserving normal data structure while spreading anomalous data. Figure~\ref{fig:vis_dtdp} (c) visualizes the DM-P energy gradient field at $ k = 1 $, with arrows guiding normal data toward high-density regions and anomalous data along the diagonal in varied directions, maintaining separation. Figure~\ref{fig:vis_dtdp} (d) illustrates data trajectories: a blue trajectory from a normal region converges to the high-density area, an orange trajectory from an anomalous region remains stationary, and a green trajectory from another anomalous region near normal data shifts toward the anomalous zone. 

\end{APPENDICES}

%%%%%%%%%%%%%%%%%
\end{document}